\documentclass[letterpaper]{article} 
\usepackage{aaai25}  
\usepackage{times}  
\usepackage{helvet}  
\usepackage{courier}  
\usepackage[hyphens]{url}  
\usepackage{graphicx} 
\usepackage{natbib}  
\usepackage{caption} 
\usepackage{multirow}
\usepackage{booktabs}
\usepackage{multicol}
\usepackage{subcaption}
\usepackage{colortbl}
\usepackage{color,xcolor}
\usepackage{amsmath}
\usepackage{tcolorbox} 
\usepackage[utf8]{inputenc} 
\usepackage[T1]{fontenc}    
\usepackage{CJKutf8}        

\usepackage{natbib}  
\usepackage{caption} 
\usepackage{algorithm}
\usepackage{algorithmic}
\usepackage{bbding}
\usepackage{soul}

\usepackage{newfloat}
\usepackage{listings}
\DeclareCaptionStyle{ruled}{labelfont=normalfont,labelsep=colon,strut=off} 
\floatstyle{ruled}
\newfloat{listing}{tb}{lst}{}
\floatname{listing}{Listing}
\usepackage{bibentry}
\definecolor{mygray}{gray}{0.9}
\definecolor{newgreen}{RGB}{78, 173, 102}
\newcommand{\worse}[2]{{#1 \textcolor{purple}{\small{-#2}}}}
\newcommand{\better}[2]{\cellcolor{white}{#1  \textcolor{newgreen}{\scriptsize{+#2}}}}
\newcommand{\gap}[2]{\cellcolor{white}{#1  \textcolor{purple}{\scriptsize{-#2}}}}
\newcommand{\bfbetter}[2]{\cellcolor{white}{\textbf{#1}  \textcolor{newgreen}{\small{+#2}}}}
\newcommand{\bfbetterr}[2]{\cellcolor{white}{#1  \textcolor{newgreen}{\small{+#2}}}}

\title{Cross-Lingual Text-Rich Visual Comprehension: An Information Theory Perspective}
\author{
    Xinmiao Yu\textsuperscript{\rm 1}, Xiaocheng Feng\textsuperscript{\rm 1}\thanks{Corresponding author}, Yun Li\textsuperscript{\rm 1}, Minghui Liao\textsuperscript{\rm 2}, Ya-Qi Yu\textsuperscript{\rm 2}, Xiachong Feng\textsuperscript{\rm 3}, Weihong Zhong\textsuperscript{\rm 1}, Ruihan Chen\textsuperscript{\rm 1}, Mengkang Hu\textsuperscript{\rm 3}, Jihao Wu\textsuperscript{\rm 2}, Dandan Tu\textsuperscript{\rm 2}, Duyu Tang\textsuperscript{\rm 2},  Bing Qin\textsuperscript{\rm 1}\footnotemark[1]
}

\affiliations{
    \textsuperscript{\rm 1}Harbin Institute of Technology\\
    \textsuperscript{\rm 2}Huawei Inc.\\
    \textsuperscript{\rm 3}The University of Hong Kong \\
    \{xmyu, xcfeng, yunli, whzhong, rhchen\}@ir.hit.edu.cn, fengxc@hku.hk, mkhu@connect.hku.hk, \\
   \{liaominghui1, yuyaqi5, wujihao, tangduyu, tudandan\}@huawei.com
}

\usepackage{bibentry}

\begin{document}

\maketitle

\begin{abstract}
Recent Large Vision-Language Models (LVLMs) have shown promising reasoning capabilities on text-rich images from charts, tables, and documents.
However, the abundant text within such images may increase the model's sensitivity to language. 
This raises the need to evaluate LVLM performance on cross-lingual text-rich visual inputs, where the language in the image differs from the language of the instructions.
To address this, we introduce \textbf{XT-VQA} (\textbf{Cross}-Lingual \textbf{Text}-Rich \textbf{V}isual \textbf{Q}uestion \textbf{A}nswering), a benchmark designed to assess how LVLMs handle language inconsistency between image text and questions.
XT-VQA integrates five existing text-rich VQA datasets and a newly collected dataset, XPaperQA, covering diverse scenarios that require faithful recognition and comprehension of visual information despite language inconsistency. Our evaluation of prominent LVLMs on XT-VQA reveals a significant drop in performance for cross-lingual scenarios, even for models with multilingual capabilities. A mutual information analysis suggests that this performance gap stems from cross-lingual questions failing to adequately activate relevant visual information. To mitigate this issue, we propose \textbf{MVCL-MI} (\textbf{M}aximization of \textbf{V}ision-Language \textbf{C}ross-\textbf{L}ingual \textbf{M}utual \textbf{I}nformation), where a visual-text cross-lingual alignment is built by maximizing mutual information between the model's outputs and visual information. This is achieved by distilling knowledge from monolingual to cross-lingual settings through KL divergence minimization, where monolingual output logits serve as a teacher. Experimental results on the XT-VQA demonstrate that MVCL-MI effectively reduces the visual-text cross-lingual performance disparity while preserving the inherent capabilities of LVLMs, shedding new light on the potential practice for improving LVLMs. Codes are available at: https://github.com/Stardust-y/XTVQA.git



\end{abstract}

%

\section{Introduction}

Large Vision-Language Models (LVLMs) have achieved significant advancements in domains such as mathematical reasoning~\cite{lu2023mathvista}, multimodal search~\cite{yang2024embodied} and embodied intelligence~\cite{mu2024embodiedgpt}. Notably, their robust multimodal capabilities demonstrate superiority in handling text-rich scenarios, leading to various applications, including document processing~\cite{luo2024layoutllm}, free-form web auto-manipulation~\cite{niu2024screenagentvisionlanguagemodeldriven}, and scene text understanding~\cite{yu2024texthawk,ye2023ureaderuniversalocrfreevisuallysituated}. These works meticulously explore the ability of LVLMs to recognize, process, and analyze multimodal information based on instructions, yielding notable progress.

\begin{figure}[t]
\centering
\includegraphics[width=1.0\columnwidth]{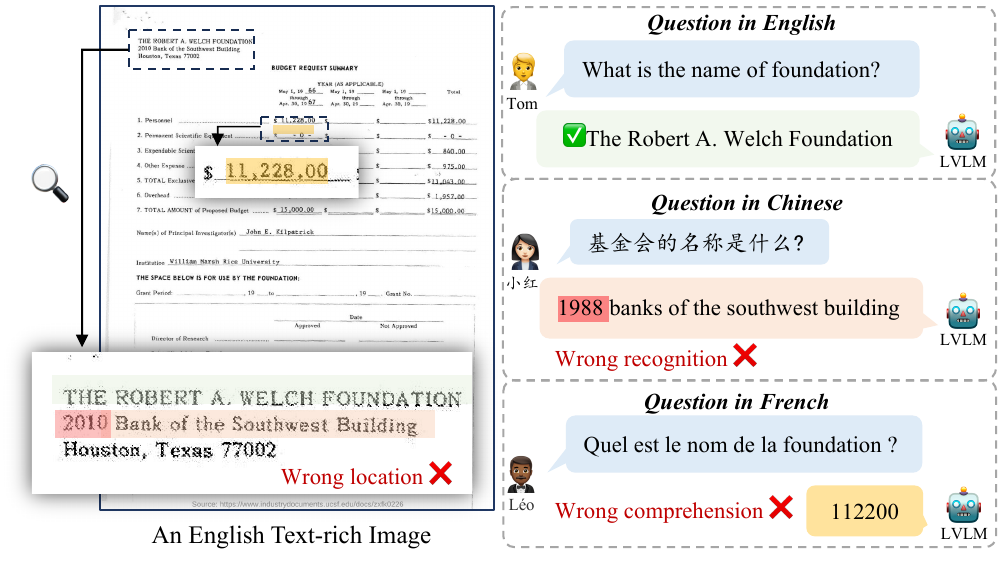} 
\caption{An example of the LVLM answering unfaithfully when questions were posed in languages different from those in the image. The LVLM made unfaithful recognition and comprehension of Chinese and French while answering correctly with English questions. Reveals the challenge of cross-lingual visual comprehension.}
\label{fig2}
\end{figure}


However, current research on text-rich visual comprehension primarily focuses on monolingual settings, largely neglecting the performance of LVLMs in cross-lingual scenarios where the instruction language differs from the textual language in the visual content, (see Fig 1.) This gap limits many real-world applications. For example, in a \textit{foreign airport with signs in unfamiliar language}, the ability to query an LVLM \textit{in your native language} for assistance would be invaluable. As globalization accelerates, cross-lingual scenarios will become increasingly common across domains such as healthcare~\cite{wan2024med}, law~\cite{guha2024legalbench}, and science~\cite{lu2022learn}. Investigating the cross-lingual instruction-following capabilities of LVLMs~\cite{hinck2024llava} is therefore essential. To address this, our work systematically explores the task of cross-lingual text-rich visual comprehension by tackling three key scientific questions.


First, we address the question: \textit{“Does the cross-lingual scenario impact the text-rich visual comprehension capabilities of LVLMs?”} To answer this, we construct the XT-VQA (\textbf{Cross}-Lingual \textbf{Text}-Rich \textbf{V}isual \textbf{Q}uestion \textbf{A}nswering) benchmark to overcome the challenge of data scarcity. XT-VQA integrates multiple existing VQA datasets~\cite{mathew2021docvqa,masry2022chartqa,singh2019vqamodelsread,mishra2019ocr} and introduces the newly curated XPaperQA dataset, which focuses on bilingual academic papers. Designed to study cross-lingual text-rich visual comprehension, XT-VQA covers diverse visual information types, including charts, scene text, and documents. XPaperQA, a key component of XT-VQA, contains 4,436 question-answer pairs generated using the advanced Gemini-Pro model, with rigorous filtering and quality review processes ensuring high data quality. Notably, XPaperQA addresses the scarcity of non-English images in existing datasets. Experimental results on XT-VQA reveal that while LVLMs demonstrate multilingual capabilities, they face significant difficulties in cross-lingual text-rich visual comprehension, with performance dropping by 32.6\%.

Next, we address the second question: \textit{``What causes the performance decline of LVLMs in cross-lingual text-rich visual comprehension scenarios?''} Inspired by prior work leveraging information theory to analyze performance gaps~\cite{farquhar2024detecting}, we examine the performance drop on XT-VQA from an information-theory perspective. Since answers in XT-VQA are typically embedded in textual form within images, effective comprehension of visual information is crucial for LVLMs to perform well. To quantify the role of visual information across languages, we analyze the mutual information between the model output and the input image. Our analysis reveals a strong correlation between accuracy and mutual information, suggesting that increasing mutual information between the visual and language components could mitigate the cross-lingual performance gap.

Finally, we address the third question: \textit{``How can we mitigate this gap while retaining monolingual capability?''} To this end, we propose \textbf{MVCL-MI} (\textbf{M}aximize \textbf{V}ision-Language \textbf{C}ross-\textbf{L}ingual \textbf{M}utual \textbf{I}nformation), a method designed to enhance the activation of visual information in LVLMs. MVCL-MI improves cross-lingual performance on XT-VQA while preserving monolingual capabilities by leveraging cross-lingual distillation to maximize mutual information between visual and language modalities across different languages. We evaluate MVCL-MI on the XT-VQA benchmark, comparing it with existing LVLMs. Experimental results show that MVCL-MI effectively reduces the performance gap in cross-lingual settings while maintaining strong monolingual performance. Ablation studies further confirm that the improvements in accuracy and fidelity stem from enhanced mutual information across modalities and languages.

\section{Related Works}

\textbf{Text-Rich Multimodal Understanding}
Text-rich multimodal understanding requires VLMs' abilities to recognize, understand, and reason over the text content contained in images \cite{mathew2021docvqa, masry2022chartqa,mishra2019ocr, singh2019vqamodelsread}. Many works try to improve text-rich visual comprehension. CLIPPO \cite{tschannen2023clippo} further improves the CLIP \cite{radford2021learningtransferablevisualmodels} by training with image and rendered text pair alignment. Pix2Struct \cite{lee2023pix2structscreenshotparsingpretraining} trains a powerful end2end model to convert text-rich screenshots into structural HTML code.  As the development of instruction fine-tuning in LLM \cite{brown2020languagemodelsfewshotlearners, touvron2023llamaopenefficientfoundation}, LVLM uses the projector to align visual tokens to text tokens and then does visual instruction tuning based on the LLM backbone \cite{liu2023visualinstructiontuning, bai2023qwenvlversatilevisionlanguagemodel, dai2023instructblipgeneralpurposevisionlanguagemodels}.  Works \cite{li2024monkey, yu2024texthawk} further enriches the instruction tuning dataset with OCR data results in OCR-related task performance rise remarkably.

\noindent\textbf{Cross-lingual in Multimodal}
Cross-lingual research is a key area in natural language processing, covering tasks such as cross-lingual information retrieval, question answering, and summarization~\cite{thakur2024leveragingllmssynthesizingtraining,wang2023zeroshotcrosslingualsummarizationlarge,chen2023revisitingcrosslingualsummarizationcorpusbased,huang2024surveylargelanguagemodels}.
While prior studies have assessed the multilingual capabilities of LVLMs~\cite{schneider2024mmathbf5diversebenchmark,wang2024seaevalmultilingualfoundationmodels}, including training models in specific languages such as Arabic~\cite{andersland2024amharicllamallavamultimodal} and English-Korean-Chinese trilingual models~\cite{shin2024xllavaoptimizingbilinguallarge}, their ability to handle cross-lingual tasks in visual contexts remains underexplored. While MTVQA~\cite{tang2024mtvqa} investigates same-language visual-text alignment in multilingual settings, our work uniquely focuses on cross-lingual inconsistencies in visual comprehension, specifically targeting real-world applications such as interpreting foreign signs.

\noindent \textbf{Information theory in Multimodal}
Information theory interrelates with deep learning tightly. \cite{tishby2015deep} employs information bottleneck as the theoretical framework for analyzing deep learning. Decoding approaches that leverage mutual information scores have demonstrated their usefulness across various scenarios~\cite{li2016mutual}. For instance, they have proven beneficial in zero-shot settings \cite{holtzman-etal-2021-surface} or when aiming to promote diversity and relevance in neural dialogue models \cite{li-etal-2016-diversity,takayama-arase-2019-relevant} 
Mutual information has been used in alleviating hallucinations in language models \cite{xiao2021hallucination}.
\cite{nandwani2023pointwisemutualinformationbased}Use conditional pointwise mutual information as score to quantify the faithfulness of models' response.

\vspace{-2mm}
\section{XT-VQA Benchmark}
\begin{figure*}[t]
\centering
\includegraphics[width=1.0\textwidth]{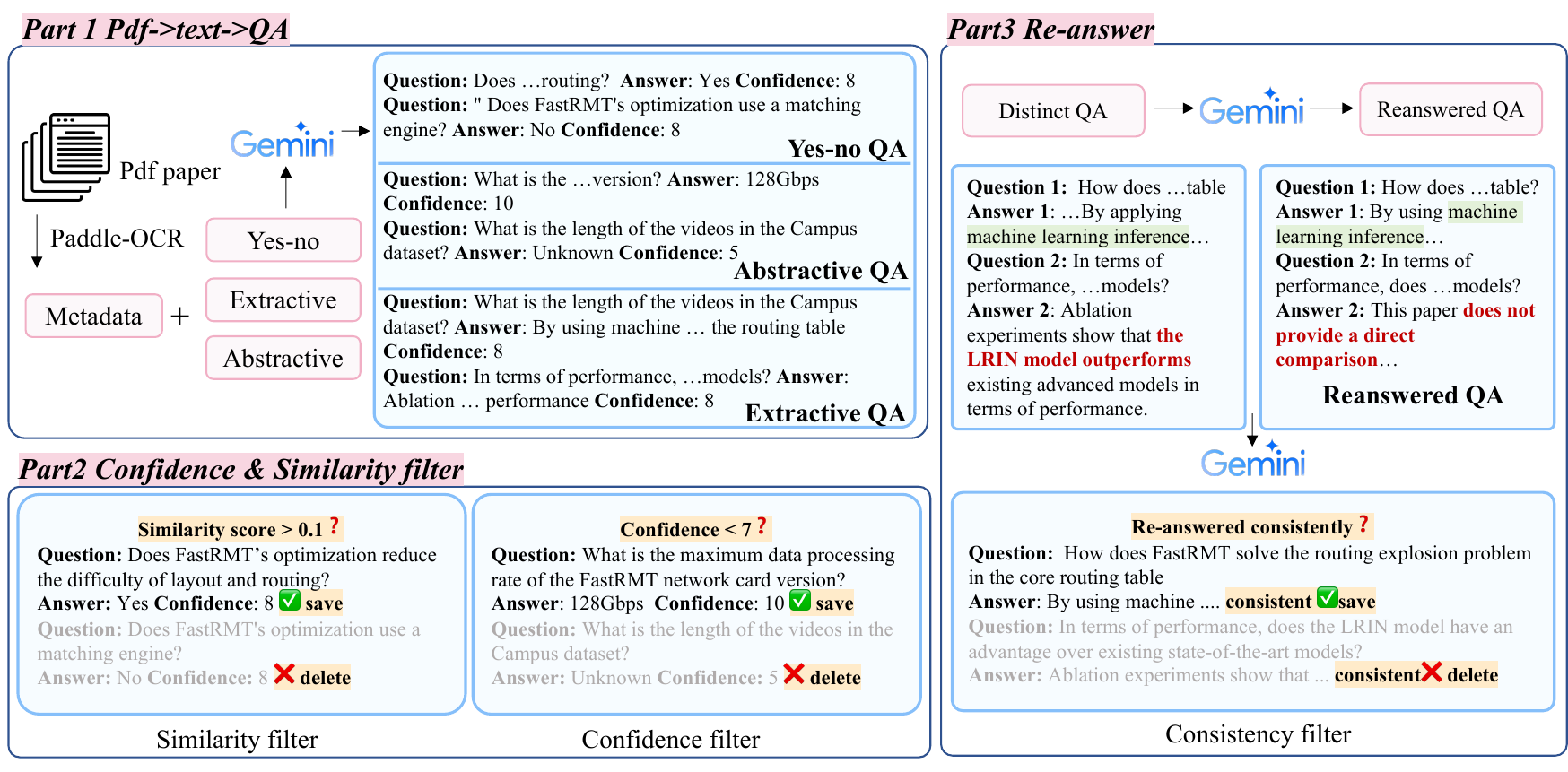} 
\caption{The XPaperQA dataset construction pipeline consists of three parts: (1) Converting PDF papers into metadata using PaddleOCR and generating three QA types via Gemini. (2) Filtering QA pairs with similarity scores $>0.1$ or confidence scores $<7$ to retain distinct pairs. (3) Re-answering the distinct QA pairs through Gemini and discarding inconsistent responses.}
\vspace{-10pt}
\label{fig3}
\end{figure*}

\subsection{Problem Formulation}
Formally, a cross-lingual text-rich question-answer pair can be represented as a text-rich image $I$ containing text in a source language $L^{src}$, a question $Q$ in target language $L^{tgt}$, where $L^{tgt}\neq L^{src}$. The goal is to accurately predict the answer $A$ to the question $Q$, by effectively leveraging the visual and textual information present in image $I$, despite the language mismatch between $L^{src}$ and $L^{tgt}$.
\vspace{-5pt}
\subsection{Dataset Construction}
\vspace{-2pt}

Our dataset construction consists of two parts. First, to evaluate the performance gap of LVLMs across different scenarios—charts, documents, and scene text—we expand four established text-rich benchmarks into three languages. Second, since existing text-rich datasets primarily contain images with English text, it is unclear whether the performance gap arises from cross-language interference or the abundance of English multimodal training data. To address this, we developed XPaperVQA, a dataset containing text-rich images in both Chinese and English. Below, we detail the dataset construction process.
\subsubsection{Multilingual Text-rich VQA extension}
\vspace{-3pt}
We use Google Translate\footnote{https://translate.google.com/?sl=en\&tl=zh-CN\&op=translate} to extend existing text-rich Visual Question Answering datasets into multiple languages: English, Chinese, and French. To improve robustness and reduce translation biases, we apply back-translation and calculate BERT sentence similarity between the original and back-translated questions. Questions with similarities below a predefined threshold are manually corrected.

\begin{equation}
\begin{split}
    \theta = \sum_{(i,j)\in WordPairs} \max(cosSim(BERT_{emb}(tokX_i), \\
    BERT_{emb}(tokY_j)))
\end{split}
\end{equation}
Here, \(cosSim\) denotes the cosine similarity between the BERT embeddings of token \(i\) from the original sentence (\(tokX_i\)) and token \(j\) from the translated sentence (\(tokY_j\)).

For English papers, we reconstruct the QASPER dataset~\cite{dasigi2021datasetinformationseekingquestionsanswers}, which contains 5,049 questions across 1,585 Natural Language Processing papers, categorized into three types: extractive, abstractive, and yes-no. To adapt this text-only dataset for cross-lingual text-rich QA, we use PyMuPDF\footnote{https://pymupdf.readthedocs.io/en/latest/} to automatically extract structural metadata from the PDF files. For each question, we locate the page containing evidence to answer it. If such a page exists, we save it as a document image along with the QA pair; otherwise, we discard the pair. This filtering process results in 1,536 VQA pairs. For Chinese papers, we develop an automatic QA generation pipeline using papers collected from an authoritative Chinese computer science journal\footnote{http://jcip.cipsc.org.cn/CN/home}.

\begin{table*}[htbp]\small\centering
\setlength\tabcolsep{3pt}
\centering
\caption{LVLMs performance on XT-VQA. Accuracy of question in source language $L^{src}$ are \textbf{bold}. \gap{29.1}{-33.6} indicates accuracy decrease compared to queries in source language $L^{src}$. The $*$ notes that auxiliary OCR tokens are used. \underline{Underline} are used to mark the highest accuracy among LVLMs. For closed-sourced models, testing is conducted on a subset only.}
\vspace{-1mm}
\begin{tabular}{l l l l l l l l l l l l l l }

\toprule
 Model & \multicolumn{3}{c}{\textbf{OCRVQA}} &  \multicolumn{3}{c}{\textbf{Text-VQA}}  & \multicolumn{3}{c}{\textbf{ChartVQA}}  &  \multicolumn{3}{c}{\textbf{DocVQA}}   \\ 
  &\multicolumn{1}{c}{\centering en} & \multicolumn{1}{c}{\centering zh} & \multicolumn{1}{c}{\centering fr} & \multicolumn{1}{c}{\centering en} & \multicolumn{1}{c}{\centering zh} & \multicolumn{1}{c}{\centering fr} & \multicolumn{1}{c}{\centering en} & \multicolumn{1}{c}{\centering zh} & \multicolumn{1}{c}{\centering fr} & \multicolumn{1}{c}{\centering en} & \multicolumn{1}{c}{\centering zh} & \multicolumn{1}{c}{\centering fr} \\ \midrule
  \textit{Open-sourced} \\ 
 LLaVA-v1.5-13b &\textbf{62.7*} & \gap{29.1*}{33.6}& \gap{31.5*}{31.2} & \textbf{61.2*} & \gap{5.9*}{55.3} & \gap{11.8*}{49.4} & \textbf{11.1} & \gap{5.8}{5.3} & \gap{7.3}{3.8} & \textbf{2.8} & \gap{1.3}{1.5} &  \gap{1.2}{1.6} &\\
 LLaVA-v1.6-34b &\textbf{64.8} &\gap{48.1}{16.7} & \gap{39.8}{25.0} & \textbf{64.9} &\gap{54.5}{10.4} & \gap{26.7}{38.2} & \textbf{52.4} & \gap{33.7}{18.7} & \gap{31.8}{20.6}  &\textbf{78.2} &\gap{61.4}{16.8} & \gap{64.0}{14.2} \\
 InstructBLIP & \textbf{24.4}&\gap{10.9}{13.5}& \gap{17.8}{6.6}& \textbf{50.3*}& \gap{34.5*}{15.8} & \gap{37.1*}{13.2} & \textbf{29.7} & \gap{21.6}{8.1} & \gap{19.8}{9.9} & \textbf{5.2}&\gap{3.1}{2.1} & \gap{4.0}{1.2} \\
 mPlug-Owl2 & \underline{\textbf{70.7}} & \underline{\gap{65.0}{5.7}} & \underline{\gap{65.2}{5.5}} &\textbf{54.3} & \gap{45.4}{8.9} & \gap{46.7}{7.6} & \textbf{44.3} & \gap{23.2}{21.1} & \gap{19.3}{25.0} &  \textbf{28.7} & \gap{21.0}{7.7} & \gap{20.7}{8.0} \\ 
 Qwen-VL-Chat & \textbf{65.6}& \gap{36.1}{29.5}& \gap{32.3}{33.3} & \textbf{61.6} & \gap{35.1}{26.5} & \gap{28.3}{33.3} & \textbf{57.3} & \gap{45.9}{11.4} & \gap{38.8}{18.5} & \textbf{59.1} & \gap{26.9}{32.2} & \gap{30.3}{28.8} \\
  Monkey & \textbf{70.4} & \gap{46.3}{24.1} & \gap{48.9}{21.5} & \textbf{61.6} & \gap{33.8}{27.8} & \gap{35.8}{25.8} & \textbf{64.6} & \gap{55.8}{8.8}& \gap{52.9}{11.7}& \textbf{65.9} & \gap{51.7}{14.2} & \gap{49.8}{16.1} \\
 Cog-VLM & \textbf{70.5} & \gap{63.8}{6.7} & \gap{61.9}{8.6} & \underline{\textbf{78.9}} & \underline{\gap{66.0}{12.9}} & \underline{\gap{66.1}{12.8}} & \textbf{57.6} & \gap{47.5}{10.1} & \gap{48.9}{8.7} & \textbf{65.4} & \gap{42.9}{22.5} & \gap{45.5}{19.9} \\
 MiniCPM-V & \textbf{69.5}& \gap{46.9}{22.6}& \gap{55.9}{13.6} & \textbf{76.6} & \gap{63.5}{13.1} & \gap{52.7}{23.9} & \underline{\textbf{73.0}} & \underline{\gap{62.9}{10.1}}& \underline{\gap{63.2}{9.8}}& \underline{\textbf{84.9}} & \underline{\gap{71.6}{13.3}} & \underline{\gap{74.6}{10.3}} \\ 
 
 \midrule
 \textit{Closed-sourced} \\
 GPT-4o & \textbf{52.3} & \gap{46.8}{5.5} & \gap{46.8}{5.5} & \textbf{72.6} & \underline{\gap{66.9}{5.7}} & \gap{65.4}{7.2} & \underline{\textbf{69.1}} & \underline{\gap{64.6}{4.5}} & \underline{\gap{64.3}{4.8}} & \textbf{74.7} & \gap{69.3}{5.4} & \gap{68.2}{6.5}\\
 Gemini-1.5-flash & \underline{\textbf{55.9}} & \underline{\gap{49.5}{6.4}} & \underline{\gap{49.5}{6.4}} & \underline{\textbf{72.7}} & \gap{64.0}{8.7} & \underline{\gap{70.7}{2.0}} & \underline{\textbf{69.1}} & \gap{59.3}{9.8} & \gap{63.1}{6.0} & \textbf{76.1} & \gap{69.8}{6.3} & \gap{70.4}{5.7}\\
  \bottomrule

\end{tabular}

\label{tab:example_selection}

\end{table*}

As shown in Figure 2, we split each paper's PDF into pages and use PaddleOCR\footnote{https://github.com/PaddlePaddle/PaddleOCR/tree/main} to extract text content. The extracted text, combined with a QA generation prompt, is input into Gemini. Following QASPER~\cite{dasigi2021datasetinformationseekingquestionsanswers}, we design three academic QA prompt types: yes-no, extractive, and abstractive. To ensure faithfulness and diversity, we apply strict filters. For accuracy, the model self-rates answers on a confidence scale of 1-10, discarding pairs with scores below $7$. For diversity, we remove pairs with Jaccard similarity exceeding $0.1$:

\begin{equation}
    J(A,B)=\frac{|A\cap B|}{|A\cup B|}
\end{equation}

Finally, to further improve the robustness, we ask Gemini to re-answer the generated question, we provide Gemini with the edit distance as a reference to detect the answer consistency between the successive answers. At last, we obtain 3,870 QA pairs with 1,039 paper images as shown in ~\ref{tab:reason}.
\begin{equation}
    d(a_1,a_2)=\min\left\{ insert, delete, replace\right\}
\end{equation}
\vspace{-5mm}

\begin{table}[ht]
\setlength{\abovecaptionskip}{-0.4pt}
\caption{Data statistics of XPaperQA}
\setlength{\tabcolsep}{2pt}
\begin{tabular}{l c c c c  }

\toprule
  & abstractive & extractive & yes-no &  image num   \\ \midrule
origin & 8,289 & 12,241 & 14,359  & 1,199  \\ 
$+$confidence filter & 8,112 & 12,040 & 14,261  & 1,199\\ 
$+$similarity filter & 1,374 & 1,530 & 1,687  & 1,074 \\
$+$consistency filter & 1,369 & 1,501 & 1,586 & 1,072 \\ \midrule
final & 1,369 & 1,501 & 1,000 & 1,039 \\ 

\bottomrule 
\end{tabular}
\label{tab:reason}
\end{table}

To ensure the quality and reliability of the XPaperQA dataset, we conducted a rigorous manual evaluation. We randomly sampled 100 questions from both the English and Chinese subsets and verified the correctness of their corresponding answers. The evaluation showed an accuracy of 87\%, demonstrating the effectiveness of our question-answering pipeline in generating high-quality pairs.

With its bilingual question-answer pairs across document scenarios, XPaperQA provides a valuable benchmark for evaluating cross-lingual multimodal understanding. It enables researchers to address challenges arising from linguistic variations between visual and textual modalities and to develop potential solutions.

\subsection{Evaluation Metrics}
With a focus on measuring the faithfulness of answers under cross-lingual instructions, we do not care about the language of the answer as long as it is correct. We uniformly translate the answer to the source language $L^{src}$ in the image. We use the F1 score to measure accuracy on XPaperQA.

\subsection{Evaluation of LVLMs on XT-VQA}
\subsubsection{Experimental Setup}
We use the respective prompt set by LVLM to get its best performance and set the temperature to the default value in the model implementation. OCR tokens were provided if the model required them by default.

We benchmark 8 open-source and 2 closed-source LVLMs on XT-VQA, reporting results separately for extended datasets and the newly collected XPaperQA. Details of models and datasets are in the Appendix. Table 2 shows LVLM performance on XT-VQA, which evaluates their ability to address language inconsistencies between image text and questions—a key challenge for text-rich data like charts, tables, and documents. The benchmark analysis reveals the following findings:

\noindent\textbf{Cross-lingual questions do produce a performance decline among the eight LVLMs.} Although LVLMs have achieved promising accuracy conditioned on English instructions, the overall average performance of these 8 LVLMs decreased by 32.5\% in Chinese and 32.6\% in French. In particular, TextVQA decreased most at 34.5\% in Chinese and 40.4\% in French, the DocVQA follows next, decreased at 33.6\% and 30.3\% separately in Chinese and French. The OCRVQA has a gap of 33.0\% in Chinese and 29.5\% in French, while performance decreases 27.9\% and 30.3\% separately on ChartQA.

\noindent\textbf{Involving multilingual data during training exhibits relative consistent cross-lingual performance.} After calculation, compared to LLaVA-v1.6-34b 24.7\% decline, Cog-VLM receives an overall 13.8\% decline conditioned on Chinese queries, while their monolingual performance is close to each other. MiniCPM-V has a 14.4\% decrease conditioned on French queries, which is also better than the 24.5\% decline of LLaVA-v1.6.34b. We suppose it is because Cog-VLM and MiniCPM-V utilize more multilingual instruction during fine-tuning.


\section{Mutual Information Analysis}
This section analyzes the performance of LVLMs on XT-VQA from an information theory perspective. We first show how we employ mutual information to examine their cross-lingual transfer capabilities. Subsequently, we present the insights derived from our mutual information analysis.

\subsection{Large Vision-Language Model Architecture}
An LVLM is typically composed of a vision encoder $E(\cdot)$, a projector $f(\cdot)$, and a Large Language Model (LLM) backbone $p_\theta$ parameterized by $\theta$. The model takes image input $I$ with a text sequence $x=[x_1,...,x_n]$ as the instruction together to generate a corresponding sequences $y=[y_1,...y_n]$. The image was first encoded by vision encoder as $E_v=E(I)$ and then projected and tokenized to the text embedding space by projector $f(E(v))$ as a sequence of visual tokens $v=[v_1, ..., v_n]$. As a Markov process, the conditional probability distribution $p_\theta(y|v,x)$ can be decomposed as
\begin{equation}
    p_\theta(y|v,x)=\prod^m_{i=1}p_\theta(y_i|v,x,y_{<i}).
\end{equation}
\vspace{-4pt}

\vspace{-8pt}
\subsection{Mutual Information in LVLM}
\vspace{-3pt}



For our cross-lingual text-rich question-answering task, we have a question $Q$ in specific language and the outputs $Y\in\mathcal{Y}$ from LVLM, and the given image $I\in\mathcal{I}$ is tokenized as $V\in\mathcal{V}$.   Given their joint distribution based on question $p(y,v|q)$, the relevance of outputs and image token $v$ is defined as the mutual information $I(Y;V|Q)$, where $V$ implicitly determines the distribution of $Y$. We want to analyze the mutual information between the outputs $Y$ and the text-rich image tokens $V$ conditioned on the cross-lingual question $Q$, formulated as
\begin{equation}
\begin{aligned}
    I(Y|V,Q) &= H(Y|Q) - H(Y|V,Q) \\&= -\sum_{|\mathcal{Y}|} P(y|V)\log P(y|V) \\&-\sum_{|\mathcal{Y}|} P(Y|V,Q)\log P(Y|V,Q).
\end{aligned}
\end{equation}
\vspace{-9pt}

Here, $H(Y|Q)$ represents the unconditional entropy of the output distribution which is invisible of the referenced image tokens $I$, while $H(Y|V,Q)$ represents the conditional entropy of the output distribution given both the image and question tokens.

Directly calculating the entropy $H(Y|V,Q)$ on the entire sentence distribution is computationally intractable due to the exponential growth of the vocabulary size $|W|$ with respect to the sequence length $l$. However, as a Markov process~\cite{jelinek1985markov}, the probability distribution between the tokens $p(y_i|y_{i<i})$ that make $Y$ is independent, we can decompose the entropy as:
\vspace{-6pt}
\begin{equation}
\begin{aligned}
    H(Y|V,Q) &= H(y_1,..y_i..,y_n|V,Q)=\sum_i^{|Y|} H(y_i|V,Q,y_{<i}) \\&= \sum_i^{|Y|} p_\theta(y_i|V,Q,y_{<i})\log p_\theta(y_i|V,Q,y_{<i}).
\end{aligned}
\end{equation}
\vspace{-4pt}

Note that $H(Y|Q)$ represents the unconditional case where the LVLM cannot see the referenced image $I$, meaning $V=\phi$. Since pure text input without images may cause unexpected effects on the LVLM distribution, we replace the image-free unconditional entropy $H(Y|Q)$ with a Gaussian noise-augmented image $I_{\epsilon} = \epsilon + I, \epsilon \sim \mathcal{N}(\mu, \sigma)$, tokenized as $V_{\epsilon}$, to ensure the stability of the final distribution. Based on the ability of heavy noise to corrupt visual information, we assume an equivalence between adding noise to the image and having no image: $H(Y|Q) = H(Y|V=\phi, Q) \approx H(Y|V_{\epsilon}, Q)$. This assumption follows visual contrastive decoding~\cite{leng2024mitigating}, where Gaussian noise approximates the unconditional distribution by reducing the influence of visual information and making outputs rely more on linguistic priors.

Finally, the mutual information is calculated as:
\begin{equation}
\begin{aligned}
    I(Y;V|Q)&=H(Y|V=\phi,Q)-H(Y|V,Q)\\&=\sum_i^{|Y|} p_\theta(y_i|V_\epsilon,Q)\log p_\theta(y_i|V_\epsilon,Q)\\&-\sum_i^{|Y|} p_\theta(y_i|V,Q)\log p_\theta(y_i|V,Q).
\end{aligned}
\end{equation}
That's how we utilize mutual information to measure the extent of activation between the input image and LVLM outputs, conditioned on queries in different languages.  A higher $I(Y; V|Q)$, suggests a stronger correlation between the image and the outputs under the condition of a given question, which aids in answers in faithfulness.
\begin{figure*}[t]
\centering
\includegraphics[width=1\textwidth]{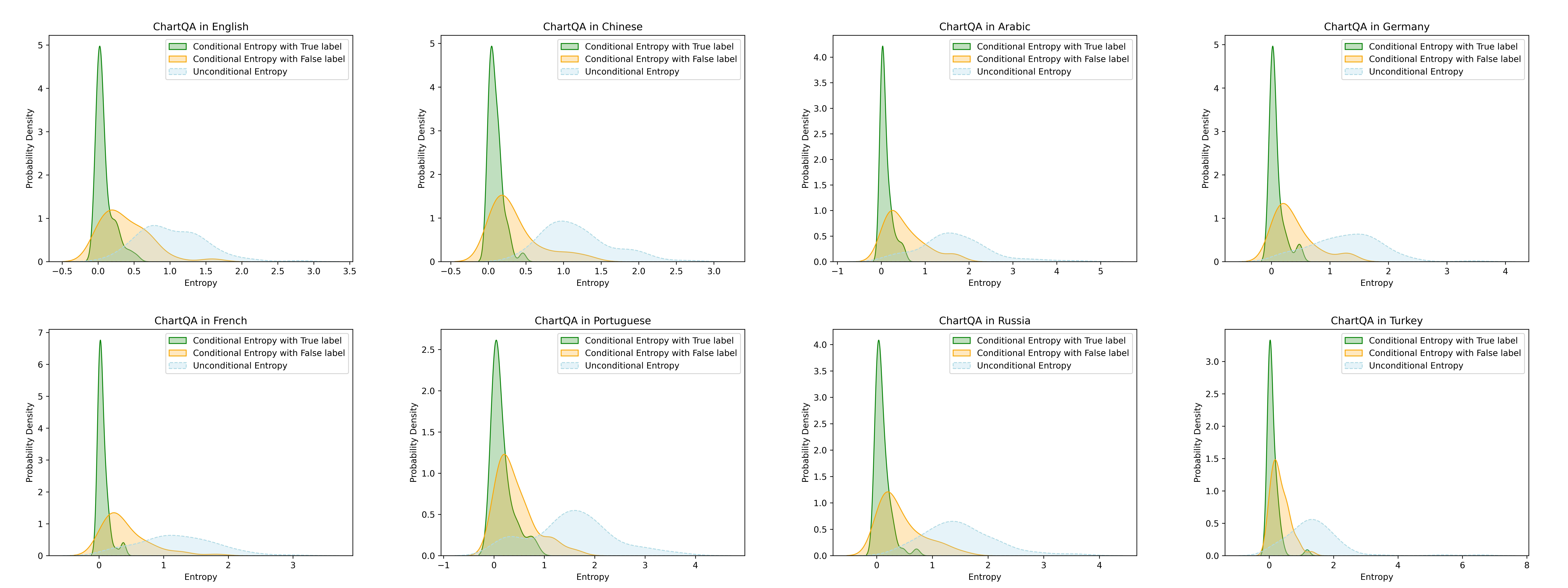} 
\caption{The entropy distribution of 100 randomly selected examples on the ChartQA dataset in 8 different languages, where the vertical axis represents probability density and the horizontal axis represents the numerical value of entropy. In all 8 languages, the mean and variance of the conditional entropy distribution for correct examples (represented in green) are significantly lower than those for incorrect examples (represented in yellow).
}
\vspace{-10pt}
\label{fig1}
\end{figure*}

\begin{figure}[t]
\centering
\includegraphics[width=0.85\columnwidth]{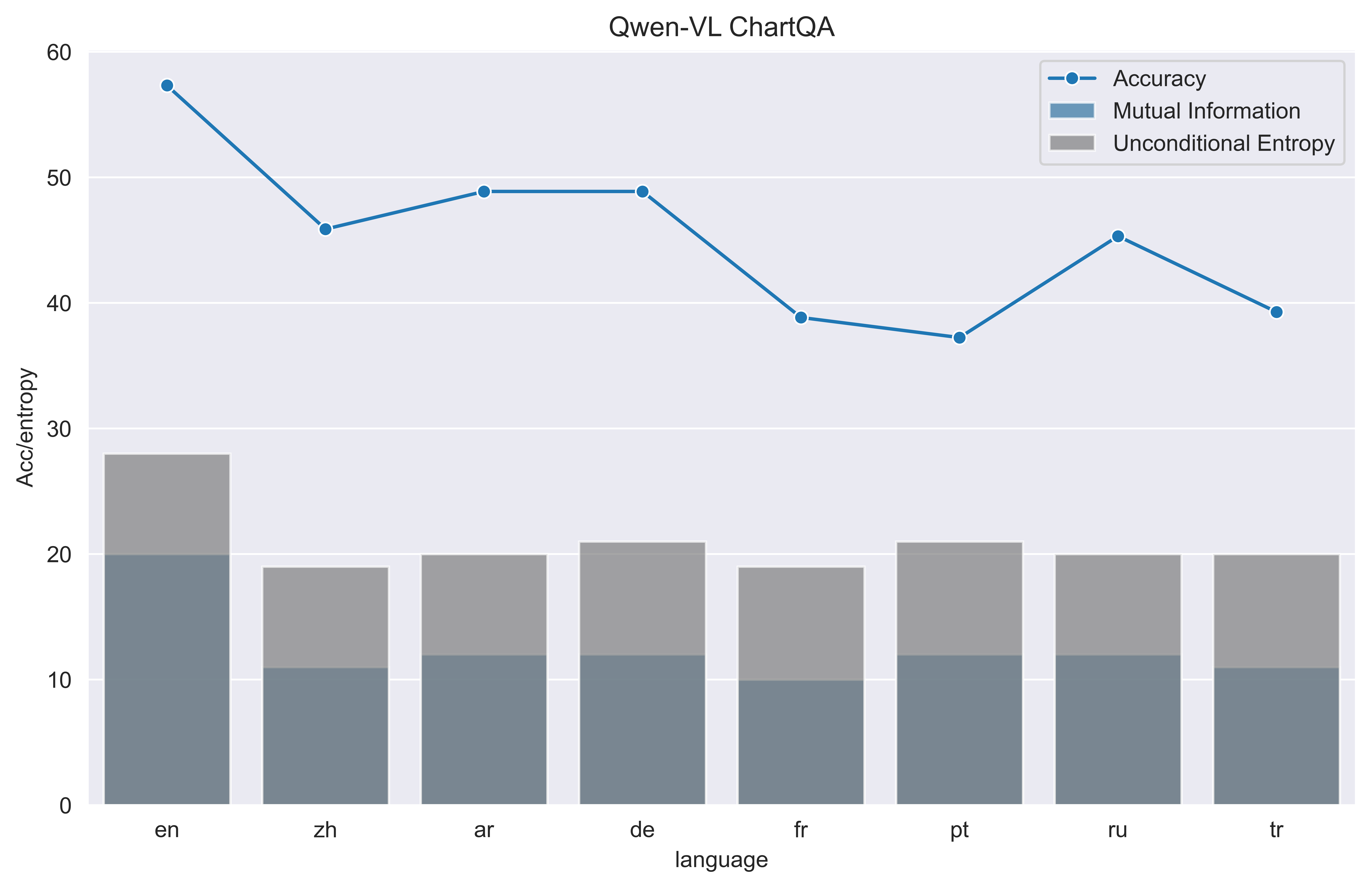} 
\caption{Statistics of accuracy and mutual information over 8 different languages on ChartQA dataset. Query in English (same in image text language) performs best, while all other languages have decreased to some extent. Reflects a correlation of accuracy and mutual information. }
\vspace{-15pt}
\label{figure2}
\end{figure}

\subsection{Mutual Information Analysis across Languages}

We randomly selected 100 examples from the ChartQA dataset.  After analyzing instructions in eight different languages, we arrived at two clear conclusions: 

\noindent\textbf{1. Entropy as a measure of uncertainty:} As depicted in Figure 4, while the unconditional entropy $H(Y|V_\epsilon,Q)$ appears random, the overall distribution of conditional entropy $H(Y|V,Q)$ for correct examples is significantly lower than for incorrect examples.  This reveals that the correct examples have higher certainty than incorrect examples.

\noindent\textbf{2. Correlation between accuracy and mutual information:} we illustrate the accuracy and mutual information of 8 different languages in Figure 3, even state-of-the-art LVLMs like Qwen-VL-Chat, which have been fine-tuned on multilingual data, display a noticeable performance disparity in cross-lingual contexts. The variation in mutual information with instructions in different languages indicates how much the visual information is activated by cross-language instructions. At last, we found a strong correlation between accuracy and mutual information, where questions in the same language as the document provide more mutual information in the answer, i.e., $I(X;Y|Q^{src}) > I(X;Y|Q^{others})$.

\section{Methodology}
In this section, we introduce \textbf{MVCL-MI} (\textbf{M}aximize \textbf{V}ision-Language \textbf{C}ross-\textbf{L}ingual \textbf{M}utual \textbf{I}nformation), to mitigate the cross-lingual performance gap on XT-VQA. 

Based on the analysis, our goal is to maximize the mutual information $I^{src} = I(Y;V|Q^{src})$ between outputs and images $V$ containing text in source language $L^{src}$ conditioned on the question in target language $L^{tgt}$ while retaining mutual information $I^{src}=I(Y;V|Q^{src})$ conditioned on $L^{src}$
\begin{equation}
\begin{aligned}
    I^{src}&=H(Y|V_\epsilon, Q^{src})-H(Y|V,Q^{src})\\
    I^{tgt}&=H(Y|V_\epsilon, Q^{tgt})-H(Y|V,Q^{tgt})
\end{aligned}
\end{equation}
As analyzed in Figure 4, the distribution of unconditional entropy is not affected by language of question $Q$, so that $H(Y|V_\epsilon,Q^{src})$ and $H(Y|V_\epsilon, Q^{tgt})$ is close to each other. To increase $I^{tgt}$ while retaining $I^{src}$, we only need to minimize $H(Y|V,Q^{tgt})$.


However, directly minimizing the entropy will lead to LVLM learning a shortcut to a sharp distribution of output logits. Instead, we use the output logits based of $Q^{src}$ as a teacher, distilling the knowledge to target language $Q^{tgt}$ by minimize the KL divergence~\cite{kullback1951information} between the distribution of $p_\theta(y^{tgt}|V,Q^{tgt})$, represented as $P^{tgt}_\theta$ and $p_\theta(y^{src}|V,Q^{src})$, represented as $P^{src}_\theta$.
We add this to training objective across languages as $\mathcal{L}_\text{KL}$.
\begin{equation}
    \mathcal{D}_\text{KL}(P^{tgt}_\theta\Vert P^{src}_\theta)=\sum^N_{i=1}p_\theta(y_i|V,Q^{tgt})\cdot(\log\frac{p_\theta(y_i|V,Q^{tgt})}{p_\theta(y_i|V,Q^{src})}).
\end{equation}


\definecolor{lightgreen}{rgb}{0.9, 0.95, 0.9}
\definecolor{lightorange}{rgb}{1, 0.9, 0.8}
\sethlcolor{lightgreen}
\begin{table*}[h]\small\centering
\setlength\tabcolsep{2pt}

\caption{LVLMs performance on XT-VQA subset XPaperQA. The best performance over tested models is marked as \underline{Underline}. Performance gap compared to $L_{src}$ is indicated as \gap{12.3}{3.9}. $\Delta$ indicates the changes compare with MiniCPM-V.}
\vspace{-0.3cm}
\begin{tabular}{lllllllllllllllll}
\toprule
 & \multicolumn{8}{c}{\textbf{XPaperQA-en}} &  \multicolumn{8}{c}{\textbf{XPaperQA-zh}}   \\
 & \multicolumn{2}{c}{extractive} & \multicolumn{2}{c}{abstractive} & \multicolumn{2}{c}{yes-no} & \multicolumn{2}{c}{overall} & \multicolumn{2}{c}{extractive} & \multicolumn{2}{c}{abstractive} & \multicolumn{2}{c}{yes-no} & \multicolumn{2}{c}{overall}\\ 
 Model &en & zh &en&zh & en&zh&en&zh & zh &en&zh & en&zh&en&zh &en \\ \midrule 
 \textit{Open-Sourced}\\
LLaVA-v1.5-13b & \textbf{9.0}&\gap{5.9}{3.1}& \textbf{12.3}&\gap{8.4}{3.9}&\textbf{50.8}&\gap{11.3}{39.5}&\textbf{14.1}&\gap{6.9}{7.2} & \textbf{8.9} & \gap{3.9}{5.0} & \textbf{16.2} & \gap{11.5}{4.7} &\textbf{68.5} & \gap{58.5}{10.0} & \textbf{27.1} & \gap{20.9}{6.2}  \\
LLaVA-v1.6-34b &\textbf{18.3}&\gap{12.0}{6.3}&\textbf{23.4}&\gap{12.8}{10.6}&\underline{\textbf{74.2}}&\gap{57.4} {16.8}&\underline{\textbf{26.8}}&\gap{16.9}{9.9} & \textbf{18.9} & \gap{7.5}{11.4} & \textbf{24.7} & \gap{10.2}{14.5} & \textbf{73.2} & \gap{66.9}{6.3} &\textbf{35.2} & \gap{23.9}{11.3}  \\
InstructBLIP &  \textbf{8.1} & \gap{1.8}{6.3} & \textbf{15.4} & \gap{12.9}{2.5} & \textbf{56.2} & \gap{50.3}{5.9} & \textbf{11.3} & \gap{7.5}{3.8}& \textbf{4.8} & \better{5.2}{0.4} & \textbf{9.9} & \gap{6.4}{3.5}& \textbf{66.1} &\gap{52.6}{13.5} & \textbf{22.6} & \gap{17.9}{4.7} \\
mPlug-Owl2 & \textbf{11.2} & \gap{9.7}{1.5} & \textbf{16.3} & \gap{12.1}{4.2} &\textbf{63.3} & \underline{\gap{61.7}{1.6}} & \textbf{17.6} & \gap{15.6}{2.0} & \textbf{14.3} & \gap{2.5}{11.8} & \textbf{5.5} & \gap{2.2}{3.3}& \textbf{68.6} & \gap{63.9}{4.7} & \textbf{24.9} &\gap{18.2}{6.7}\\
Qwen-VL-Chat & \textbf{13.1} & \gap{9.9}{3.2} & \textbf{15.1} &\gap{11.7}{3.4} &\textbf{50.0} &\gap{32.7}{17.3} &\textbf{16.8} & \gap{12.4}{4.4}&\textbf{9.8}&\gap{6.6}{3.2}&\textbf{23.8}&\underline{\gap{17.7}{6.1}}&\textbf{69.3}&\gap{61.6}{7.7} &\textbf{30.6} & \gap{25.1}{5.5} \\
Monkey &\textbf{20.6}&\gap{15.1}{5.5}&\textbf{16.2}&\gap{6.1}{10.1}&\textbf{49.2} & \gap{45.8}{3.4}&\textbf{21.5}&\gap{15.4}{6.1}&\textbf{17.4}&\gap{12.7}{4.7}&\textbf{18.4}&\gap{14.2}{4.2} &\textbf{73.1}&\gap{60.8}{12.3}& \textbf{32.2} & \gap{25.7}{6.5} \\
Cog-VLM &\textbf{16.2}&\gap{12.4}{3.8}&\underline{\textbf{22.2}}& \underline{\gap{14.8}{7.4}}&\textbf{47.5}&\better{53.4}{5.9} &\textbf{20.4}&\gap{16.8}{3.6}& \textbf{21.3} & \gap{9.9}{11.4} & \textbf{25.1} & \gap{16.2}{8.9} & \textbf{74.5} & \gap{69.3}{5.2} & \textbf{36.5} & \gap{27.7}{8.8}  \\
MiniCPM-V&\underline{\textbf{25.8}}&\underline{\gap{22.3}{3.5}}&\textbf{14.7}&\gap{12.7}{2.0}&\textbf{56.5} &\gap{37.4}{19.1}&\textbf{25.3}&\underline{\gap{20.4}{4.9}}&\underline{\textbf{36.6}}&\underline{\gap{18.5}{18.1}}&\underline{\textbf{34.7}}&\gap{16.9}{17.8}&\underline{\textbf{77.6}}&\underline{\gap{75.4}{2.2}}& \underline{\textbf{46.1}}& \underline{\gap{32.6}{13.5}} \\ \midrule
\textit{ours} \\
MVCL-MI (8B) & \textbf{25.9} & \gap{22.5}{3.4} & \textbf{16.0} & \gap{14.1}{1.9} & \textbf{60.4} & \gap{52.3}{8.1} & \textbf{25.9} & \gap{22.5}{3.4} & \textbf{36.2} & \gap{23.9}{12.3} & \textbf{37.2} & \gap{23.2}{14.0} & \textbf{81.3} & \gap{79.0}{2.3} & \textbf{48.2} & \gap{37.9}{10.3} \\

$\Delta$ & \hl{$\uparrow$0.1} & \hl{$\uparrow$0.2} & \hl{$\uparrow$1.3} & \hl{$\uparrow$1.4} & \hl{$\uparrow$3.9}&\hl{$\uparrow$15.1} &\hl{$\uparrow$0.6} & \hl{$\uparrow$1.9} & \sethlcolor{lightorange}\hl{$\downarrow$0.4}  & \sethlcolor{lightgreen}\hl{$\uparrow$5.4} & \hl{$\uparrow$2.5} & \hl{$\uparrow$6.3} & \hl{$\uparrow$3.7} & \hl{$\uparrow$3.6} & \hl{$\uparrow$2.1} &\hl{$\uparrow$5.3} \\
\bottomrule

\end{tabular}
\label{tab:main_result}
\vspace{-6pt}

\end{table*}

\noindent\textbf{Training Objectives} Finally, our training objective is
\begin{equation}
\begin{aligned}
    \mathcal{L}&=\mathcal{L}_{\text{CE}}(y^{src-src},\hat{y}^{src})+\mathcal{L}_{\text{CE}}(y^{tgt-src},\hat{y}^{src})\\
    &+\mathcal{L}_{\text{CE}}(y^{src-tgt},\hat{y}^{tgt})+\mathcal{L}_{\text{CE}}(y^{tgt-tgt},\hat{y}^{tgt}) \\
    &+\alpha\mathcal{L}_{\text{KL}}(P^{src-tgt}||P^{tgt-tgt})+\beta\mathcal{L}_{\text{KL}}(P^{tgt-src}||P^{src-src}).
\end{aligned}
\label{trainingobj}
\end{equation}
Here, $y^{src-tgt}$ represents the output logits of LVLM $p_\theta(y^{src}|Q^{tgt},V)$ queried in source language and asks the model answer in the target language, same goes for others.

\(\mathcal{L}_\text{CE}\) denotes the cross-entropy loss for source and target language, which is equivalent to maximizing the likelihood of the ground truth answers $y_{src}$ and $y_{tgt}$ given the image and question in the respective languages. The terms $D_{KL}$ represent the KL divergence between the predicted distributions in different languages, with $\alpha$ and $\beta$ being hyperparameters controlling the importance of these divergence terms.

\subsection{Experiment Results}
In Table 3, we show how our proposed MVCL-MI is effective by comparing the cross-lingual gap with other LVLMs. We delve into the impact of different types of questions and different language documentation on the performance gap on XPaperQA. The further ablation study demonstrates the necessity of our training objective design.

\noindent\textbf{The cross-lingual gap exists no matter the language of the source language of images.} After calculating, the overall average performance gap of 8 LVLMs is 28.1\% in XPaperQA-en and 24.3\% in XPaperQA-zh. In the monolingual setting, LLaVA-v1.6-34b has the best performance of 26.8 in English paper, and MiniCPM performs best of 46.1 in Chinese paper. In the cross-lingual setting, MiniCPM-V performs best in both English and Chinese papers, with an accuracy of 22.5 and 32.6 respectively.

\noindent\textbf{Cross-lingual gap varies in different types of questions. } In English paper, the decrease on the 3 types of questions is 31.4\% in abstractive, 31.0\% in extractive, and 22.0\% in yes-no. In Chinese paper, the average decrease in the 3 types of questions is 44.2\% in extractive, 39.8\% in abstractive, and 11.1\% in yes-no, exhibit $Gap_\text{ext} > Gap_\text{abs} > Gap_\text{yesno}$. Since answering correctly to the abstractive and extractive questions demands higher comprehension of visual information than yes-no questions, it reflects that cross-lingual do affects the capabilities of LVLMs.

\noindent\textbf{MVCL-MI effectively mitigates the cross-lingual gap despite the language of the text in images.} Compared to the original model, our model effectively improved the performance on cross-lingual settings, with an increase of 1.9 ($\uparrow$9.3\%) on XPaperQA-en and 5.3 ($\uparrow$16.3\%) on XPaper-zh while preserving the monolingual performance.
MVCL-MI effectively narrows the performance gap from 19.4\% to 13.1\% ($\downarrow$32.5\%) in XPaper-en and 29.3\% to 21.3\% ($\downarrow$27.3\%) in XPaper-zh.




\begin{table}\small\centering

\setlength\tabcolsep{1.2pt} 
\caption{Ablation Study of MVCL-MI. \underline{Underline} indicates the original overall performance, the best was \textbf{Bold}.}
\vspace{-5pt}
\begin{tabular}{lllllllll}
\toprule
&\multicolumn{2}{c}{yes-no} &\multicolumn{2}{c}{extractive} & \multicolumn{2}{c}{abstractive} &  \multicolumn{2}{c}{overall} \\ 
Method & \multicolumn{1}{c}{\centering en} & \multicolumn{1}{c}{\centering zh} & \multicolumn{1}{c}{\centering en} & \multicolumn{1}{c}{\centering zh} & \multicolumn{1}{c}{\centering en} & \multicolumn{1}{c}{\centering zh} & \multicolumn{1}{c}{\centering en} & \multicolumn{1}{c}{\centering zh} \\ \midrule
\textit{English Paper} \\
\textit{origin} & 56.5&37.4&25.9&22.3&14.7&12.7&\underline{25.3}&\underline{20.4} \\ 
MCVL-MI &60.4 &52.3&25.9&22.5&16.0&14.1&\bfbetter{26.0}{0.7}&\bfbetter{22.5}{1.9}\\
\textit{w/o} Cross-CE & 56.6 &49.2&24.9&22.1&15.3&13.9	&\worse{25.1}{0.2}&\bfbetterr{21.7}{1.3} \\
\textit{w/o} KL-Loss &58.5&42.6&	24.4&17.8&17.6&10.8&\bfbetterr{25.3}{0.0}&\worse{19.7}{0.7} \\
\midrule
\toprule

&\multicolumn{1}{c}{\centering zh} & \multicolumn{1}{c}{\centering en} & \multicolumn{1}{c}{\centering zh} & \multicolumn{1}{c}{\centering en} & \multicolumn{1}{c}{\centering zh} & \multicolumn{1}{c}{\centering en} & \multicolumn{1}{c}{\centering zh} & \multicolumn{1}{c}{\centering en} \\\midrule
\textit{Chinese Paper} \\
origin&77.6&75.4&35.6&18.5&	34.7&16.9&\underline{46.1}&\underline{32.6} \\
MCVL-MI & 81.3 &79.0 &36.2 &23.9 &37.2 &23.2 &\bfbetter{48.2}{2.1} & \bfbetter{37.9}{5.3} \\
\textit{w/o} Cross-CE  & 77.9 &76.9 &35.4&19.9&36.6&17.1 &\bfbetterr{46.8}{0.7} &\bfbetterr{33.5}{0.9} \\ 
\textit{w/o} KL-Loss & 73.2&70.7&34.8&17.7&31.2&19.2&\worse{43.3}{2.8}&\worse{31.9}{0.7} \\

\bottomrule

\end{tabular}
\vspace{-10pt}
\end{table}

\subsection{Ablation Study}

    
We design the following ablation study to test the effectiveness of our MVCL-MI training objectives, as depicted in Tab.4.
The $\textit{w/o}$ KL-Loss setting entails removing the KL-Loss from the training objectives, shown in F.~\ref{trainingobj}. The $\textit{w/o}$ Cross-CE indicates we remove the cross-lingual Cross Entropy Loss from the training objectives F.~\ref{trainingobj}. 
It is evident that when we remove Cross-CE, the cross-lingual performance of 21.7 declines compared to MVCL-MI 22.5 in the English paper, and declines 4.4 in the Chinese paper. This result reveals the effectivity of involving cross-lingual tuning data. Removing KL-Loss even performs worse than the original model in cross-lingual settings. 
We suppose the KL divergence terms are necessary to ensure that the predicted distributions for cross-lingual scenarios ($y^{tgt-src}$ and $y^{src-tgt}$) remain close to the corresponding monolingual predictions ($y^{src-src}$ and $y^{tgt-tgt}$), which serve as anchor points. This objective is equivalent to maximizing the mutual information between the answers and input images across multiple languages while preserving the model's performance in monolingual settings.
\vspace{-3pt}

\subsubsection{Training details}
We deploy our training method on the advanced LVLM \textbf{MiniCPM-Llama3-V}. We trained on 8 A100-sxm4-80gb for 1 epoch with the default training configuration and hyperparameter setup. We report this setup in the Appendix.
\vspace{-4mm}
\section{Conclusion}

In this paper, we investigate the cross-lingual gap in text-rich visual comprehension and propose XT-VQA, a benchmark for testing LVLMs' ability to handle language inconsistencies across modalities. From an information perspective, we identify that this gap arises from insufficient activation of visual information by cross-lingual queries. To address this, we mitigate the gap by maximizing cross-lingual mutual information. Results show that MVCL-MI enables LVLMs to effectively leverage both visual and textual information, producing accurate and language-consistent answers across languages. We believe this research advances text-rich visual comprehension and enhances LVLMs' global accessibility, fostering inclusive and cross-cultural communication.

\bibliography{aaai25}

\begin{thebibliography}{45}
\providecommand{\natexlab}[1]{#1}

\bibitem[{Andersland(2024)}]{andersland2024amharicllamallavamultimodal}
Andersland, M. 2024.
\newblock Amharic LLaMA and LLaVA: Multimodal LLMs for Low Resource Languages.
\newblock arXiv:2403.06354.

\bibitem[{Bai et~al.(2023)Bai, Bai, Yang, Wang, Tan, Wang, Lin, Zhou, and Zhou}]{bai2023qwenvlversatilevisionlanguagemodel}
Bai, J.; Bai, S.; Yang, S.; Wang, S.; Tan, S.; Wang, P.; Lin, J.; Zhou, C.; and Zhou, J. 2023.
\newblock Qwen-VL: A Versatile Vision-Language Model for Understanding, Localization, Text Reading, and Beyond.
\newblock arXiv:2308.12966.

\bibitem[{Brown et~al.(2020)Brown, Mann, Ryder, Subbiah, Kaplan, Dhariwal, Neelakantan, Shyam, Sastry, Askell, Agarwal, Herbert-Voss, Krueger, Henighan, Child, Ramesh, Ziegler, Wu, Winter, Hesse, Chen, Sigler, Litwin, Gray, Chess, Clark, Berner, McCandlish, Radford, Sutskever, and Amodei}]{brown2020languagemodelsfewshotlearners}
Brown, T.~B.; Mann, B.; Ryder, N.; Subbiah, M.; Kaplan, J.; Dhariwal, P.; Neelakantan, A.; Shyam, P.; Sastry, G.; Askell, A.; Agarwal, S.; Herbert-Voss, A.; Krueger, G.; Henighan, T.; Child, R.; Ramesh, A.; Ziegler, D.~M.; Wu, J.; Winter, C.; Hesse, C.; Chen, M.; Sigler, E.; Litwin, M.; Gray, S.; Chess, B.; Clark, J.; Berner, C.; McCandlish, S.; Radford, A.; Sutskever, I.; and Amodei, D. 2020.
\newblock Language Models are Few-Shot Learners.
\newblock arXiv:2005.14165.

\bibitem[{Chen et~al.(2023)Chen, Zhang, Zhou, Bai, Wang, Zhong, Yan, Li, Li, Zhu, and Zhang}]{chen2023revisitingcrosslingualsummarizationcorpusbased}
Chen, Y.; Zhang, H.; Zhou, Y.; Bai, X.; Wang, Y.; Zhong, M.; Yan, J.; Li, Y.; Li, J.; Zhu, M.; and Zhang, Y. 2023.
\newblock Revisiting Cross-Lingual Summarization: A Corpus-based Study and A New Benchmark with Improved Annotation.
\newblock arXiv:2307.04018.

\bibitem[{Dai et~al.(2023)Dai, Li, Li, Tiong, Zhao, Wang, Li, Fung, and Hoi}]{dai2023instructblipgeneralpurposevisionlanguagemodels}
Dai, W.; Li, J.; Li, D.; Tiong, A. M.~H.; Zhao, J.; Wang, W.; Li, B.; Fung, P.; and Hoi, S. 2023.
\newblock InstructBLIP: Towards General-purpose Vision-Language Models with Instruction Tuning.
\newblock arXiv:2305.06500.

\bibitem[{Dasigi et~al.(2021)Dasigi, Lo, Beltagy, Cohan, Smith, and Gardner}]{dasigi2021datasetinformationseekingquestionsanswers}
Dasigi, P.; Lo, K.; Beltagy, I.; Cohan, A.; Smith, N.~A.; and Gardner, M. 2021.
\newblock A Dataset of Information-Seeking Questions and Answers Anchored in Research Papers.
\newblock arXiv:2105.03011.

\bibitem[{Farquhar et~al.(2024)Farquhar, Kossen, Kuhn, and Gal}]{farquhar2024detecting}
Farquhar, S.; Kossen, J.; Kuhn, L.; and Gal, Y. 2024.
\newblock Detecting hallucinations in large language models using semantic entropy.
\newblock \emph{Nature}, 630(8017): 625--630.

\bibitem[{Guha et~al.(2024)Guha, Nyarko, Ho, R{\'e}, Chilton, Chohlas-Wood, Peters, Waldon, Rockmore, Zambrano et~al.}]{guha2024legalbench}
Guha, N.; Nyarko, J.; Ho, D.; R{\'e}, C.; Chilton, A.; Chohlas-Wood, A.; Peters, A.; Waldon, B.; Rockmore, D.; Zambrano, D.; et~al. 2024.
\newblock Legalbench: A collaboratively built benchmark for measuring legal reasoning in large language models.
\newblock \emph{Advances in Neural Information Processing Systems}, 36.

\bibitem[{Hinck et~al.(2024)Hinck, Holtermann, Olson, Schneider, Yu, Bhiwandiwalla, Lauscher, Tseng, and Lal}]{hinck2024llava}
Hinck, M.; Holtermann, C.; Olson, M.~L.; Schneider, F.; Yu, S.; Bhiwandiwalla, A.; Lauscher, A.; Tseng, S.; and Lal, V. 2024.
\newblock Why do LLaVA Vision-Language Models Reply to Images in English?
\newblock \emph{arXiv preprint arXiv:2407.02333}.

\bibitem[{Holtzman et~al.(2021)Holtzman, West, Shwartz, Choi, and Zettlemoyer}]{holtzman-etal-2021-surface}
Holtzman, A.; West, P.; Shwartz, V.; Choi, Y.; and Zettlemoyer, L. 2021.
\newblock Surface Form Competition: Why the Highest Probability Answer Isn{'}t Always Right.
\newblock In Moens, M.-F.; Huang, X.; Specia, L.; and Yih, S. W.-t., eds., \emph{Proceedings of the 2021 Conference on Empirical Methods in Natural Language Processing}, 7038--7051. Online and Punta Cana, Dominican Republic: Association for Computational Linguistics.

\bibitem[{Huang et~al.(2024)Huang, Mo, Li, Li, Zhang, Yi, Mao, Liu, Xu, Xu, Nie, and Liu}]{huang2024surveylargelanguagemodels}
Huang, K.; Mo, F.; Li, H.; Li, Y.; Zhang, Y.; Yi, W.; Mao, Y.; Liu, J.; Xu, Y.; Xu, J.; Nie, J.-Y.; and Liu, Y. 2024.
\newblock A Survey on Large Language Models with Multilingualism: Recent Advances and New Frontiers.
\newblock arXiv:2405.10936.

\bibitem[{Jelinek(1985)}]{jelinek1985markov}
Jelinek, F. 1985.
\newblock Markov source modeling of text generation.
\newblock In \emph{The impact of processing techniques on communications}, 569--591. Springer.

\bibitem[{Kullback and Leibler(1951)}]{kullback1951information}
Kullback, S.; and Leibler, R.~A. 1951.
\newblock On information and sufficiency.
\newblock \emph{The annals of mathematical statistics}, 22(1): 79--86.

\bibitem[{Lee et~al.(2023)Lee, Joshi, Turc, Hu, Liu, Eisenschlos, Khandelwal, Shaw, Chang, and Toutanova}]{lee2023pix2structscreenshotparsingpretraining}
Lee, K.; Joshi, M.; Turc, I.; Hu, H.; Liu, F.; Eisenschlos, J.; Khandelwal, U.; Shaw, P.; Chang, M.-W.; and Toutanova, K. 2023.
\newblock Pix2Struct: Screenshot Parsing as Pretraining for Visual Language Understanding.
\newblock arXiv:2210.03347.

\bibitem[{Leng et~al.(2024)Leng, Zhang, Chen, Li, Lu, Miao, and Bing}]{leng2024mitigating}
Leng, S.; Zhang, H.; Chen, G.; Li, X.; Lu, S.; Miao, C.; and Bing, L. 2024.
\newblock Mitigating object hallucinations in large vision-language models through visual contrastive decoding.
\newblock In \emph{Proceedings of the IEEE/CVF Conference on Computer Vision and Pattern Recognition}, 13872--13882.

\bibitem[{Li et~al.(2016)Li, Galley, Brockett, Gao, and Dolan}]{li-etal-2016-diversity}
Li, J.; Galley, M.; Brockett, C.; Gao, J.; and Dolan, B. 2016.
\newblock A Diversity-Promoting Objective Function for Neural Conversation Models.
\newblock In Knight, K.; Nenkova, A.; and Rambow, O., eds., \emph{Proceedings of the 2016 Conference of the North {A}merican Chapter of the Association for Computational Linguistics: Human Language Technologies}, 110--119. San Diego, California: Association for Computational Linguistics.

\bibitem[{Li and Jurafsky(2016)}]{li2016mutual}
Li, J.; and Jurafsky, D. 2016.
\newblock Mutual information and diverse decoding improve neural machine translation.
\newblock \emph{arXiv preprint arXiv:1601.00372}.

\bibitem[{Li et~al.(2024)Li, Yang, Liu, Ma, Zhang, Yang, Sun, Liu, and Bai}]{li2024monkey}
Li, Z.; Yang, B.; Liu, Q.; Ma, Z.; Zhang, S.; Yang, J.; Sun, Y.; Liu, Y.; and Bai, X. 2024.
\newblock Monkey: Image resolution and text label are important things for large multi-modal models.
\newblock In \emph{Proceedings of the IEEE/CVF Conference on Computer Vision and Pattern Recognition}, 26763--26773.

\bibitem[{Liu et~al.(2023)Liu, Li, Wu, and Lee}]{liu2023visualinstructiontuning}
Liu, H.; Li, C.; Wu, Q.; and Lee, Y.~J. 2023.
\newblock Visual Instruction Tuning.
\newblock arXiv:2304.08485.

\bibitem[{Lu et~al.(2023)Lu, Bansal, Xia, Liu, Li, Hajishirzi, Cheng, Chang, Galley, and Gao}]{lu2023mathvista}
Lu, P.; Bansal, H.; Xia, T.; Liu, J.; Li, C.; Hajishirzi, H.; Cheng, H.; Chang, K.-W.; Galley, M.; and Gao, J. 2023.
\newblock Mathvista: Evaluating mathematical reasoning of foundation models in visual contexts.
\newblock \emph{arXiv preprint arXiv:2310.02255}.

\bibitem[{Lu et~al.(2022)Lu, Mishra, Xia, Qiu, Chang, Zhu, Tafjord, Clark, and Kalyan}]{lu2022learn}
Lu, P.; Mishra, S.; Xia, T.; Qiu, L.; Chang, K.-W.; Zhu, S.-C.; Tafjord, O.; Clark, P.; and Kalyan, A. 2022.
\newblock Learn to explain: Multimodal reasoning via thought chains for science question answering.
\newblock \emph{Advances in Neural Information Processing Systems}, 35: 2507--2521.

\bibitem[{Luo et~al.(2024)Luo, Shen, Zhu, Zheng, Yu, and Yao}]{luo2024layoutllm}
Luo, C.; Shen, Y.; Zhu, Z.; Zheng, Q.; Yu, Z.; and Yao, C. 2024.
\newblock LayoutLLM: Layout Instruction Tuning with Large Language Models for Document Understanding.
\newblock In \emph{Proceedings of the IEEE/CVF Conference on Computer Vision and Pattern Recognition}, 15630--15640.

\bibitem[{Masry et~al.(2022)Masry, Long, Tan, Joty, and Hoque}]{masry2022chartqa}
Masry, A.; Long, D.~X.; Tan, J.~Q.; Joty, S.; and Hoque, E. 2022.
\newblock Chartqa: A benchmark for question answering about charts with visual and logical reasoning.
\newblock \emph{arXiv preprint arXiv:2203.10244}.

\bibitem[{Mathew, Karatzas, and Jawahar(2021)}]{mathew2021docvqa}
Mathew, M.; Karatzas, D.; and Jawahar, C. 2021.
\newblock Docvqa: A dataset for vqa on document images.
\newblock In \emph{Proceedings of the IEEE/CVF winter conference on applications of computer vision}, 2200--2209.

\bibitem[{Mishra et~al.(2019)Mishra, Shekhar, Singh, and Chakraborty}]{mishra2019ocr}
Mishra, A.; Shekhar, S.; Singh, A.~K.; and Chakraborty, A. 2019.
\newblock Ocr-vqa: Visual question answering by reading text in images.
\newblock In \emph{2019 international conference on document analysis and recognition (ICDAR)}, 947--952. IEEE.

\bibitem[{Mu et~al.(2024)Mu, Zhang, Hu, Wang, Ding, Jin, Wang, Dai, Qiao, and Luo}]{mu2024embodiedgpt}
Mu, Y.; Zhang, Q.; Hu, M.; Wang, W.; Ding, M.; Jin, J.; Wang, B.; Dai, J.; Qiao, Y.; and Luo, P. 2024.
\newblock Embodiedgpt: Vision-language pre-training via embodied chain of thought.
\newblock \emph{Advances in Neural Information Processing Systems}, 36.

\bibitem[{Nandwani et~al.(2023)Nandwani, Kumar, Raghu, Joshi, and Lastras}]{nandwani2023pointwisemutualinformationbased}
Nandwani, Y.; Kumar, V.; Raghu, D.; Joshi, S.; and Lastras, L.~A. 2023.
\newblock Pointwise Mutual Information Based Metric and Decoding Strategy for Faithful Generation in Document Grounded Dialogs.
\newblock arXiv:2305.12191.

\bibitem[{Niu et~al.(2024)Niu, Li, Wang, Fu, Hu, Leng, Kong, Chang, and Wang}]{niu2024screenagentvisionlanguagemodeldriven}
Niu, R.; Li, J.; Wang, S.; Fu, Y.; Hu, X.; Leng, X.; Kong, H.; Chang, Y.; and Wang, Q. 2024.
\newblock ScreenAgent: A Vision Language Model-driven Computer Control Agent.
\newblock arXiv:2402.07945.

\bibitem[{Radford et~al.(2021)Radford, Kim, Hallacy, Ramesh, Goh, Agarwal, Sastry, Askell, Mishkin, Clark, Krueger, and Sutskever}]{radford2021learningtransferablevisualmodels}
Radford, A.; Kim, J.~W.; Hallacy, C.; Ramesh, A.; Goh, G.; Agarwal, S.; Sastry, G.; Askell, A.; Mishkin, P.; Clark, J.; Krueger, G.; and Sutskever, I. 2021.
\newblock Learning Transferable Visual Models From Natural Language Supervision.
\newblock arXiv:2103.00020.

\bibitem[{Schneider and Sitaram(2024)}]{schneider2024mmathbf5diversebenchmark}
Schneider, F.; and Sitaram, S. 2024.
\newblock M$\mathbf5$ -- A Diverse Benchmark to Assess the Performance of Large Multimodal Models Across Multilingual and Multicultural Vision-Language Tasks.
\newblock arXiv:2407.03791.

\bibitem[{Shin et~al.(2024)Shin, Lim, Won, Choi, Kim, Song, Yoo, Kim, and Lim}]{shin2024xllavaoptimizingbilinguallarge}
Shin, D.; Lim, H.; Won, I.; Choi, C.; Kim, M.; Song, S.; Yoo, H.; Kim, S.; and Lim, K. 2024.
\newblock X-LLaVA: Optimizing Bilingual Large Vision-Language Alignment.
\newblock arXiv:2403.11399.

\bibitem[{Singh et~al.(2019)Singh, Natarajan, Shah, Jiang, Chen, Batra, Parikh, and Rohrbach}]{singh2019vqamodelsread}
Singh, A.; Natarajan, V.; Shah, M.; Jiang, Y.; Chen, X.; Batra, D.; Parikh, D.; and Rohrbach, M. 2019.
\newblock Towards VQA Models That Can Read.
\newblock arXiv:1904.08920.

\bibitem[{Takayama and Arase(2019)}]{takayama-arase-2019-relevant}
Takayama, J.; and Arase, Y. 2019.
\newblock Relevant and Informative Response Generation using Pointwise Mutual Information.
\newblock In Chen, Y.-N.; Bedrax-Weiss, T.; Hakkani-Tur, D.; Kumar, A.; Lewis, M.; Luong, T.-M.; Su, P.-H.; and Wen, T.-H., eds., \emph{Proceedings of the First Workshop on NLP for Conversational AI}, 133--138. Florence, Italy: Association for Computational Linguistics.

\bibitem[{Tang et~al.(2024)Tang, Liu, Ye, Lu, Wei, Lin, Li, Mahmood, Feng, Zhao et~al.}]{tang2024mtvqa}
Tang, J.; Liu, Q.; Ye, Y.; Lu, J.; Wei, S.; Lin, C.; Li, W.; Mahmood, M. F. F.~B.; Feng, H.; Zhao, Z.; et~al. 2024.
\newblock MTVQA: Benchmarking Multilingual Text-Centric Visual Question Answering.
\newblock \emph{arXiv preprint arXiv:2405.11985}.

\bibitem[{Thakur et~al.(2024)Thakur, Ni, Ábrego, Wieting, Lin, and Cer}]{thakur2024leveragingllmssynthesizingtraining}
Thakur, N.; Ni, J.; Ábrego, G.~H.; Wieting, J.; Lin, J.; and Cer, D. 2024.
\newblock Leveraging LLMs for Synthesizing Training Data Across Many Languages in Multilingual Dense Retrieval.
\newblock arXiv:2311.05800.

\bibitem[{Tishby and Zaslavsky(2015)}]{tishby2015deep}
Tishby, N.; and Zaslavsky, N. 2015.
\newblock Deep learning and the information bottleneck principle.
\newblock In \emph{2015 ieee information theory workshop (itw)}, 1--5. IEEE.

\bibitem[{Touvron et~al.(2023)Touvron, Lavril, Izacard, Martinet, Lachaux, Lacroix, Rozière, Goyal, Hambro, Azhar, Rodriguez, Joulin, Grave, and Lample}]{touvron2023llamaopenefficientfoundation}
Touvron, H.; Lavril, T.; Izacard, G.; Martinet, X.; Lachaux, M.-A.; Lacroix, T.; Rozière, B.; Goyal, N.; Hambro, E.; Azhar, F.; Rodriguez, A.; Joulin, A.; Grave, E.; and Lample, G. 2023.
\newblock LLaMA: Open and Efficient Foundation Language Models.
\newblock arXiv:2302.13971.

\bibitem[{Tschannen, Mustafa, and Houlsby(2023)}]{tschannen2023clippo}
Tschannen, M.; Mustafa, B.; and Houlsby, N. 2023.
\newblock Clippo: Image-and-language understanding from pixels only.
\newblock In \emph{Proceedings of the IEEE/CVF Conference on Computer Vision and Pattern Recognition}, 11006--11017.

\bibitem[{Wan et~al.(2024)Wan, Liu, Zhang, Fu, Wang, Cheng, Ma, Quilodr{\'a}n-Casas, and Arcucci}]{wan2024med}
Wan, Z.; Liu, C.; Zhang, M.; Fu, J.; Wang, B.; Cheng, S.; Ma, L.; Quilodr{\'a}n-Casas, C.; and Arcucci, R. 2024.
\newblock Med-unic: Unifying cross-lingual medical vision-language pre-training by diminishing bias.
\newblock \emph{Advances in Neural Information Processing Systems}, 36.

\bibitem[{Wang et~al.(2024)Wang, Liu, Huang, Jiao, Ding, Aw, and Chen}]{wang2024seaevalmultilingualfoundationmodels}
Wang, B.; Liu, Z.; Huang, X.; Jiao, F.; Ding, Y.; Aw, A.; and Chen, N.~F. 2024.
\newblock SeaEval for Multilingual Foundation Models: From Cross-Lingual Alignment to Cultural Reasoning.
\newblock arXiv:2309.04766.

\bibitem[{Wang et~al.(2023)Wang, Liang, Meng, Zou, Li, Qu, and Zhou}]{wang2023zeroshotcrosslingualsummarizationlarge}
Wang, J.; Liang, Y.; Meng, F.; Zou, B.; Li, Z.; Qu, J.; and Zhou, J. 2023.
\newblock Zero-Shot Cross-Lingual Summarization via Large Language Models.
\newblock arXiv:2302.14229.

\bibitem[{Xiao and Wang(2021)}]{xiao2021hallucination}
Xiao, Y.; and Wang, W.~Y. 2021.
\newblock On hallucination and predictive uncertainty in conditional language generation.
\newblock \emph{arXiv preprint arXiv:2103.15025}.

\bibitem[{Yang et~al.(2024)Yang, Zhou, Li, Tao, Li, Shen, He, Jiang, and Shi}]{yang2024embodied}
Yang, Y.; Zhou, T.; Li, K.; Tao, D.; Li, L.; Shen, L.; He, X.; Jiang, J.; and Shi, Y. 2024.
\newblock Embodied multi-modal agent trained by an llm from a parallel textworld.
\newblock In \emph{Proceedings of the IEEE/CVF Conference on Computer Vision and Pattern Recognition}, 26275--26285.

\bibitem[{Ye et~al.(2023)Ye, Hu, Xu, Ye, Yan, Xu, Li, Tian, Qian, Zhang, Jin, He, Lin, and Huang}]{ye2023ureaderuniversalocrfreevisuallysituated}
Ye, J.; Hu, A.; Xu, H.; Ye, Q.; Yan, M.; Xu, G.; Li, C.; Tian, J.; Qian, Q.; Zhang, J.; Jin, Q.; He, L.; Lin, X.~A.; and Huang, F. 2023.
\newblock UReader: Universal OCR-free Visually-situated Language Understanding with Multimodal Large Language Model.
\newblock arXiv:2310.05126.

\bibitem[{Yu et~al.(2024)Yu, Liao, Wu, Liao, Zheng, and Zeng}]{yu2024texthawk}
Yu, Y.-Q.; Liao, M.; Wu, J.; Liao, Y.; Zheng, X.; and Zeng, W. 2024.
\newblock Texthawk: Exploring efficient fine-grained perception of multimodal large language models.
\newblock \emph{arXiv preprint arXiv:2404.09204}.

\end{thebibliography}

\clearpage
\section{Appendix}

\subsection{Training Details}
We introduce the setting of hyperparameters when we use MVCL-MI to fine-tune MiniCPM-V. We basically reuse the original hyperparameters setting.
\begin{table}[ht]
\centering
\setlength{\abovecaptionskip}{-0.4pt}
\caption{Training hyperparameters of MVCL-MI}
\begin{tabular}{r | l }

\toprule
Params & Value \\ \midrule
bf16 & True \\
fp16 & False \\
model\_max\_length & 2048 \\
learning\_rate & 1e-6 \\
weight\_decay & 0.1 \\
adam\_beta2 & 0.95 \\
warmup\_ratio & 0.01 \\
lr\_scheduler\_type & ``cosine'' \\ 

\bottomrule 
\end{tabular}

\end{table}

\subsection{Dataset Details}

Here's detailed information on four datasets we used to build the XT-VQA benchmark: OCRVQA, TeXT-VQA, ChartQA, and DocVQA.

\textbf{OCRVQA} is a large-scale dataset comprised of English book covers with questions about the book's name, author, dates and so on. Challenges in requiring LVLMs to have the ability of OCR and layout recognition.

\textbf{TextVQA} Pictures contained in TeXT-VQA are natural photos containing English scene text in it. TeXT-VQA requires models to read and reason about text in images to answer questions about them. Specifically, models need to incorporate a new modality of text present in the images and reason over it to answer TeXT-VQA questions.

\textbf{ChartQA} is a dataset specifically designed for answering questions about charts. The Charts contain pie charts, column charts et cetra. ChartQA is divided into two split by difficulty: human and augmented. Human split requires more reasoning and calculation ability of VLMs.

\textbf{DocQA} is a large-scale dataset of 12,767 document images of varied types and content, with over 50,000 questions and answers.
The questions defined are categorized based on their reasoning requirements, allowing us to analyze how DocVQA methods fare for different question types.

\textbf{QASPER} is a dataset for question answering on scientific research papers. It consists of 5,049 questions over 1,585 Natural Language Processing papers. Each question is written by an NLP practitioner who reads only the title and abstract of the corresponding paper, and the question seeks information present in the full text. The questions are then answered by a separate set of NLP practitioners who also provide supporting evidence to answers.

\begin{figure*}[b]
    \centering
    \begin{subfigure}[b]{0.88\textwidth}
        \includegraphics[width=\textwidth]{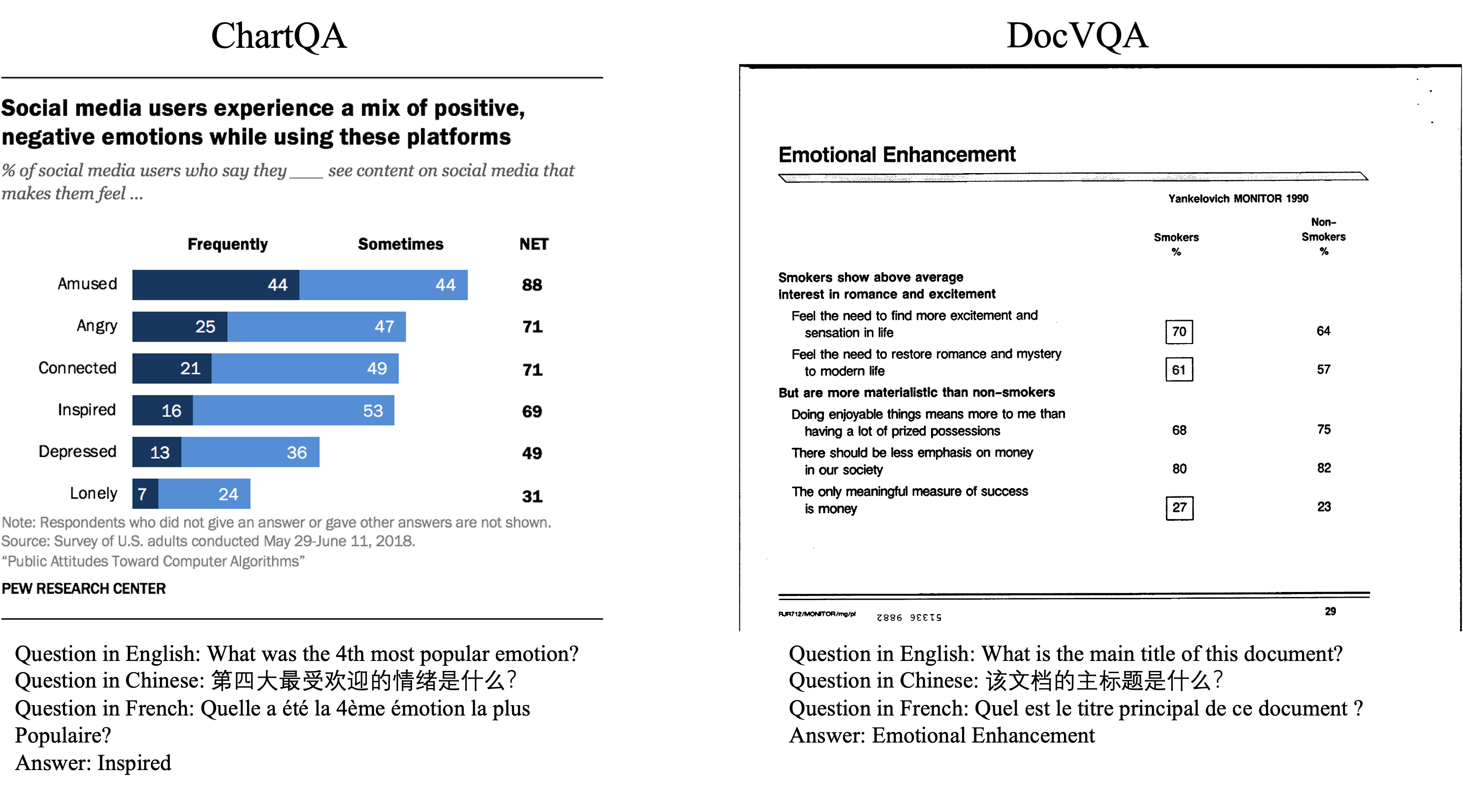}
       
    \end{subfigure}
    \hfill
    \begin{subfigure}[b]{0.88\textwidth}
        \includegraphics[width=\textwidth]{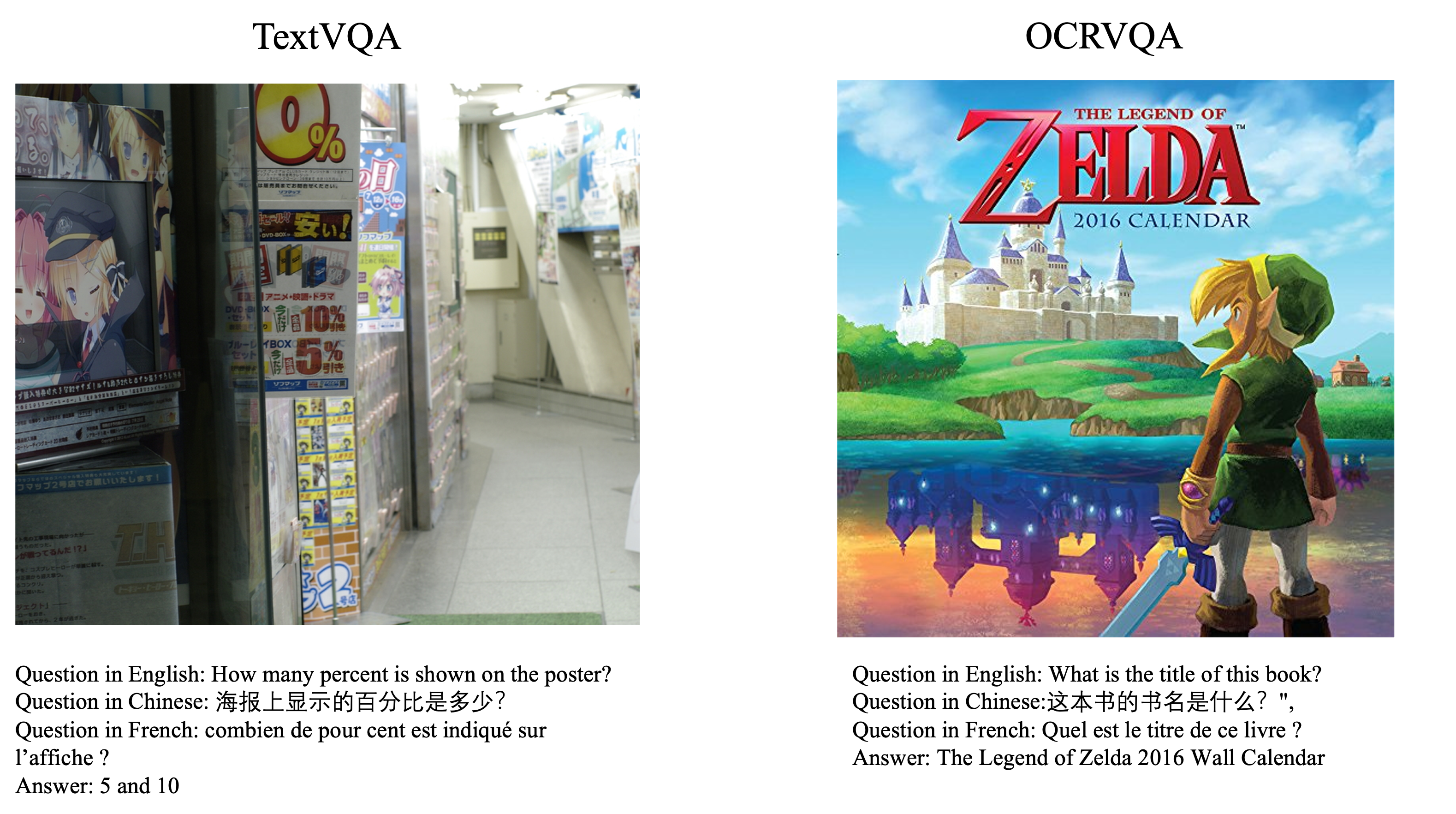}
    \end{subfigure}
    \caption{Data example from XT-VQA}
\end{figure*}

\subsection{Model Details}
We benchmark XT-VQA with the following wide-used LVLMs.

\textbf{LLaVA} is an end-to-end trained large multimodal model that is designed to understand and generate content based on both visual inputs (images) and textual instructions. It combines the capabilities of a visual encoder and a language model (Llama) to process and respond to multimodal inputs. It presents the first attempt to use language-only GPT-4 to generate multimodal language-image instruction-following data. LLaVA-Next further improved the ability to process high-resolution images. LLaVA-v1 uses Vicuna-13B-v1.3, LLaMA-2-13B-Chat and LLaMA-2-7B-Chat as base Large Language Model (LLM), CLIP-L-336px and CLIP-L as Vision Encoder.

\textbf{LLaVA-Next} LLaVA-NeXT retains the minimalist design and data efficiency of LLaVA-1.5, reusing its pretrained connector and utilizing less than 1M visual instruction tuning samples. It achieves top performance compared to open-source language-vision models like CogVLM or Yi-VL, and matches or outperforms commercial models like Gemini Pro and Qwen-VL-Plus on selected benchmarks. Notably, LLaVA-NeXT demonstrates emerging zero-shot Chinese multimodal capabilities, despite only using English data, with state-of-the-art performance on MMBench-CN. Trained on 32 GPUs for ~1 day with 1.3M samples, LLaVA-NeXT's compute and training data cost is 100-1000 times smaller than others. In addition to Vicuna-1.5 (7B and 13B), it considers various LLMs like Mistral-7B and Nous-Hermes-2-Yi-34B, with desirable properties, flexible commercial use terms, strong bilingual support, and larger language model capacity, allowing LLaVA to support a wider user base and more scenarios. The LLaVA recipe works well with various LLMs and scales smoothly up to 34B parameters.

\textbf{mPLUG-Owl-2} mPLUG-Owl-2 is an innovative training paradigm that enables large language models (LLMs) to acquire multimodal capabilities through modularized learning of a foundation LLM, a visual knowledge module, and a visual abstractor module. This modular approach supports multiple modalities and facilitates diverse unimodal and multimodal abilities through modality collaboration. It comprises a vision foundation model for encoding visual knowledge, a language foundation model, and a visual abstractor module. 

\textbf{InstructBLIP} proposes a new vision-language instruction-tuning framework using BLIP-2 models, achieving state-of-the-art zero-shot generalization performance on a wide range of vision-language tasks. Trained on 13 held-in datasets, InstructBLIP attains state-of-the-art zero-shot performance across all 13 held-out datasets, substantially outperforming BLIP-2 and larger Flamingo models. InstructBLIP uses a diverse set of instruction data to train a multimodal LLM. Specifically, It initialize training with a pre-trained BLIP-2 model consisting of an image encoder, an LLM, and a Query Transformer (Q-Former) to bridge the two. In the experiments, InstructBLIP adopt four variations of BLIP-2 with the same image encoder (ViT-g/14) but different frozen LLMs, including FlanT5-
XL (3B), FlanT5-XXL (11B), Vicuna-7B and Vicuna-13B.

\textbf{Qwen-VL-Chat} The Qwen-VL-Chat Model is a state-of-the-art Large Vision-Language Model (LVLM) developed by Alibaba Cloud. It is designed to understand and generate human-like text based on both visual and textual inputs. It starts from Qwen-LM-7B and Openclip’s ViT-bigG-1.9B. To alleviate the efficiency issues arising from long image feature sequences, Qwen-VL introduces a vision-language adapter that compresses the image features. This adapter comprises a single-layer cross-attention module initialized randomly. The module uses a group of trainable vectors (Embeddings) as query vectors and the image features from the visual encoder as keys for cross attention operations. On real-world dialog benchmarks, instruction-tuned Qwen-VL-Chat demonstrates superiority compared to existing vision-language chatbots. 

\textbf{Monkey} The Monkey model was based on Qwen-VL-Chat architecture, further finetuned with fine-grit image captioning annotated by GPT and other visual experts. Monkey has achieved impressive results on OCR-related tasks. Monkey's training conducts experiments based on the well-trained Vit-BigG and LLM from QwenVL, the pre-trained large multimodal model. Since the vision encoder has already been well-trained, Monkey proceeds directly to the instruction-tuning stage. The overall parameters for Monkey is 9.8B.

\textbf{MiniCPM-V} MiniCPM-V is a multimodal language model series for vision-language tasks, taking images and text as input to generate text outputs. The latest 2.6 version, built on SigLip-400M and Qwen2-7B with 8B parameters, shows significant performance gains over 2.5. It introduces multi-image and video understanding capabilities, outperforming models like GPT-4V across vision-language tasks. 2.6 inherits strengths like OCR, trustworthiness, and multilingualism from 2.5, while enabling real-time video understanding on end-side devices due to superior token density.

\textbf{CogVLM} CogVLM is a powerful open-source visual language model that deeply integrates vision and language through a trainable visual expert module. The 17B version, with 10B visual and 7B language parameters, supports high-resolution image understanding and multi-turn dialogues. It achieves state-of-the-art performance across 10 cross-modal benchmarks by fusing multimodal features without compromising natural language abilities. CogVLM comprises a vision transformer, adapter, large language model, and the innovative visual expert enabling deep multimodal fusion.

\onecolumn

\section{XPaperQA}
\subsection{XPaperQA English example}
\begin{figure}[ht]
\centering
\includegraphics[width=0.6\columnwidth]{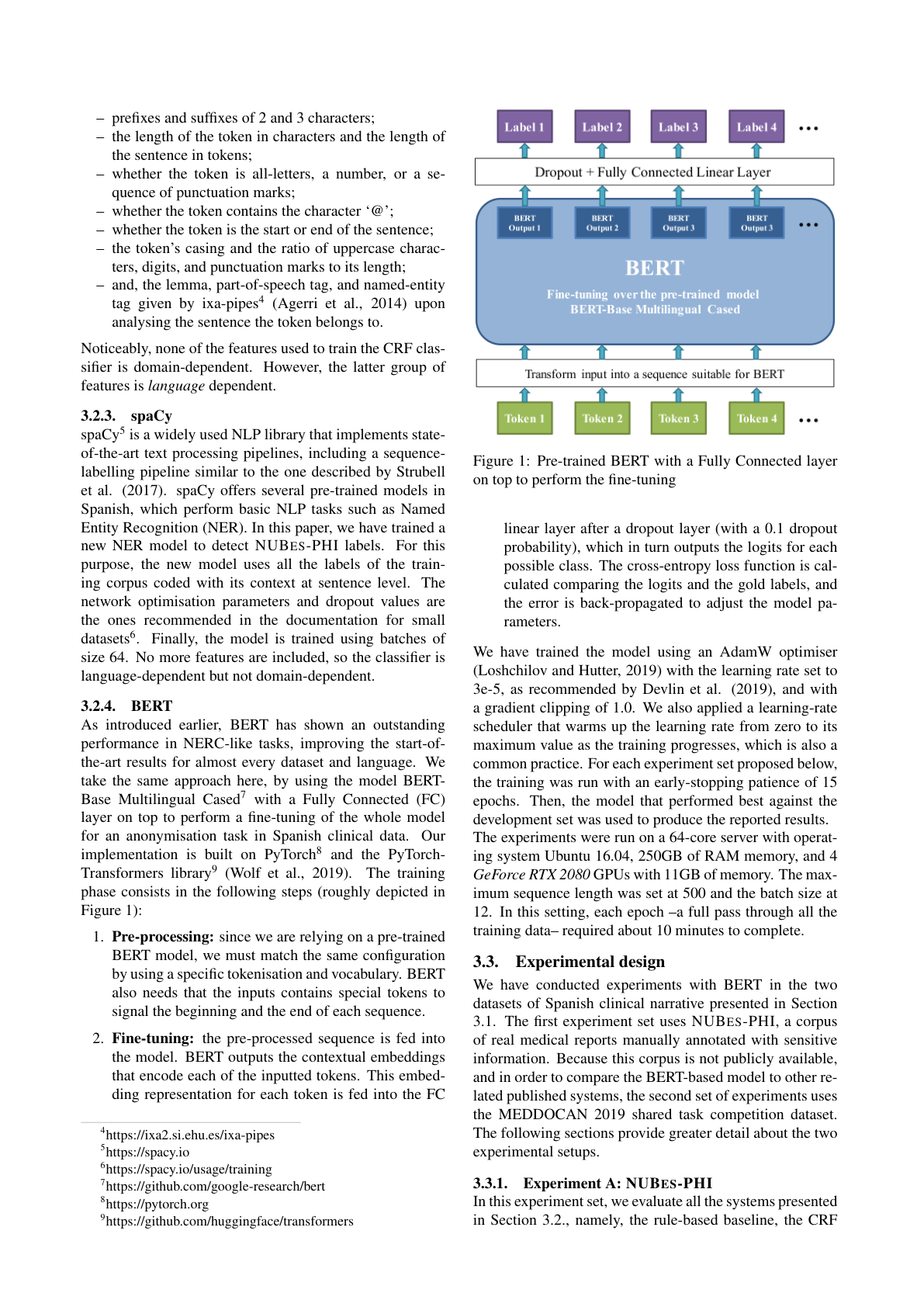} 
\caption{An example of extractive QA from XPaperQA-en dataset. }

\label{figure_example}
\end{figure}
\begin{CJK}{UTF8}{gbsn}
Question in English: What is the performance of BERT on the task?

\vspace{10pt}
Question in Chinese: BERT 在该任务上的表现如何？
\vspace{10pt}

Answer in English: BERT remains only 0.3 F1-score points behind, and would have achieved the second position among all the MEDDOCAN shared task competitors. Taking into account that only 3\% of the gold labels remain incorrectly annotated.
\vspace{10pt}

Answer in Chinese: BERT 仅落后 0.3 F1 分数点，并将在所有 MEDDOCAN 共享任务竞争对手中取得第二名。考虑到只有 3\% 的黄金标签仍然注释错误。

\end{CJK}
\clearpage
\subsection{XPaperQA Chinese example}

\begin{figure}[ht]
\centering
\includegraphics[width=0.6\columnwidth]{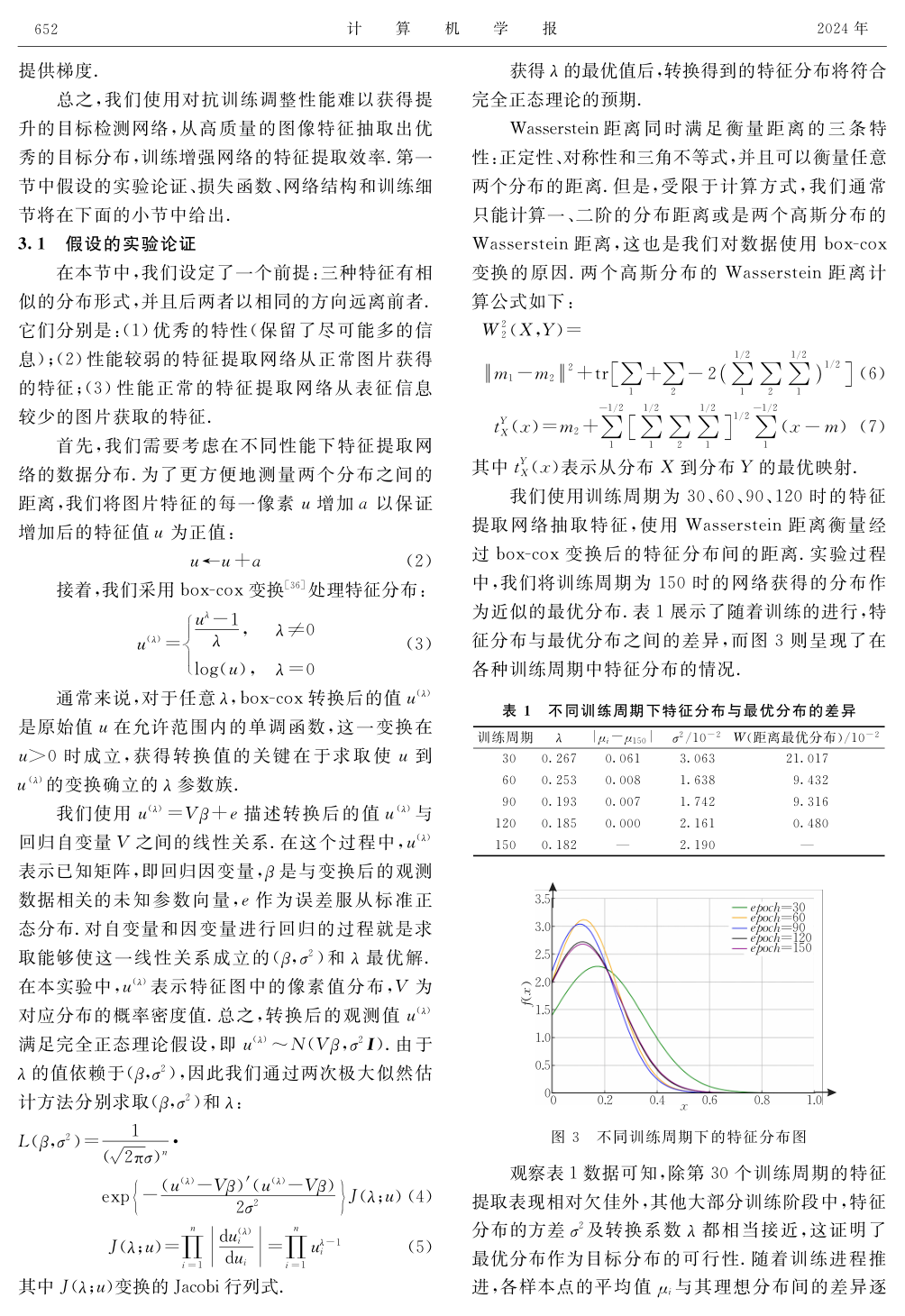} 
\caption{An example of abstractive QA from XPaperQA-zh dataset. }
\vspace{-15pt}
\label{figure_ch}
\end{figure}
\begin{CJK}{UTF8}{gbsn}
\vspace{10pt}
Question in Chinese: 在本文提出的方法中，Wasserstein距离被用来衡量什么？
\vspace{10pt}

Question in English: In the proposed method, what is Wasserstein distance used to measure?
\vspace{10pt}

Answer in Chinese: 特征分布之间的距离
\vspace{10pt}

Answer in English: The distance between feature distributions.
\vspace{10pt}

Confidence: 8

\end{CJK}
\clearpage
\subsection{Instruction Visualization}
As shown in Figure 8, we visualized the distribution of problems by sampling 200 items from Chinese and English XPaperQA datasets respectively.
\begin{figure*}[h]
    \centering
    \begin{subfigure}[b]{0.43\textwidth}
        \includegraphics[width=\textwidth]{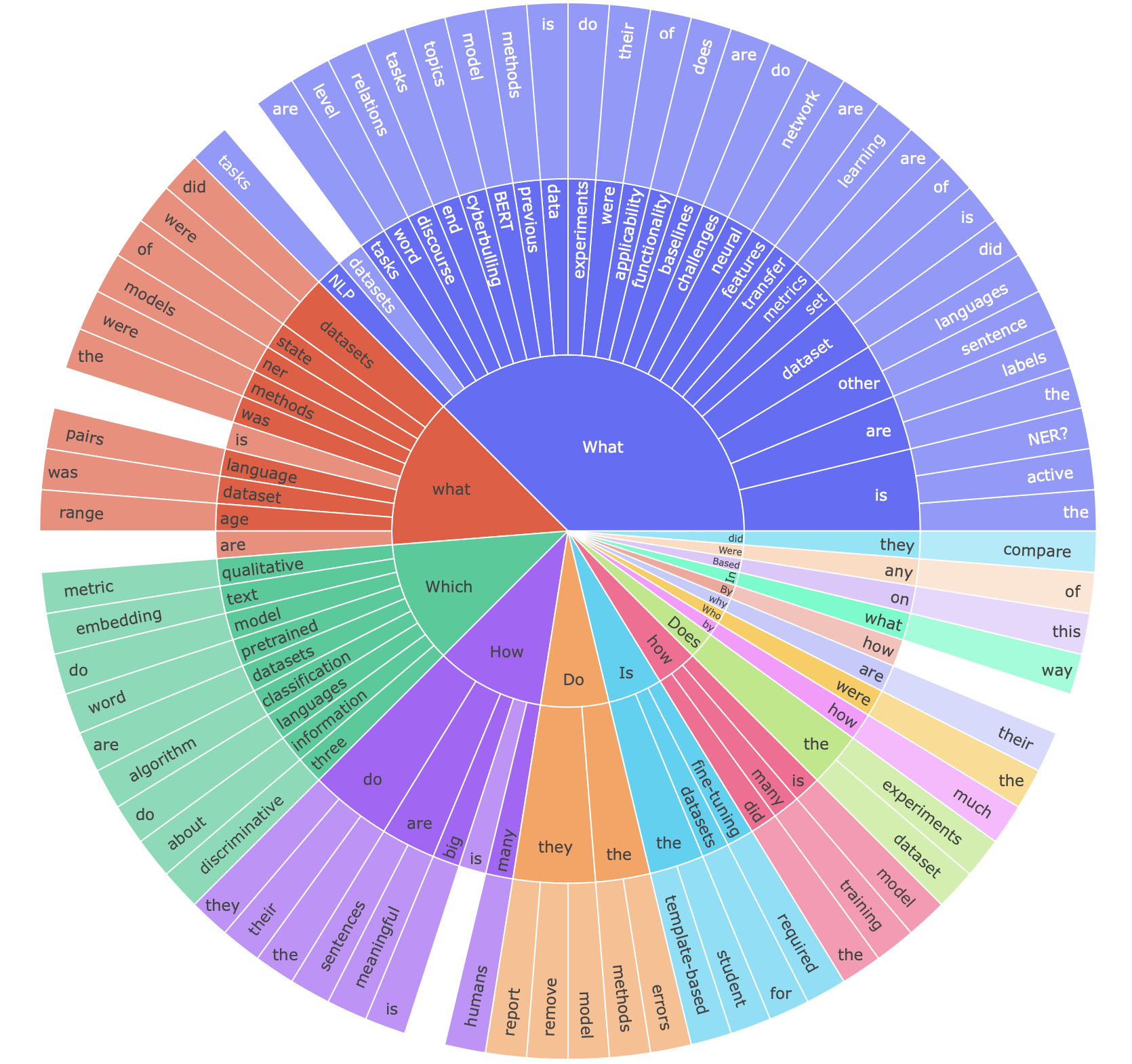}
        \caption{English paper}
    \end{subfigure}
    \hfill
    \begin{subfigure}[b]{0.42\textwidth}
        \includegraphics[width=\textwidth]{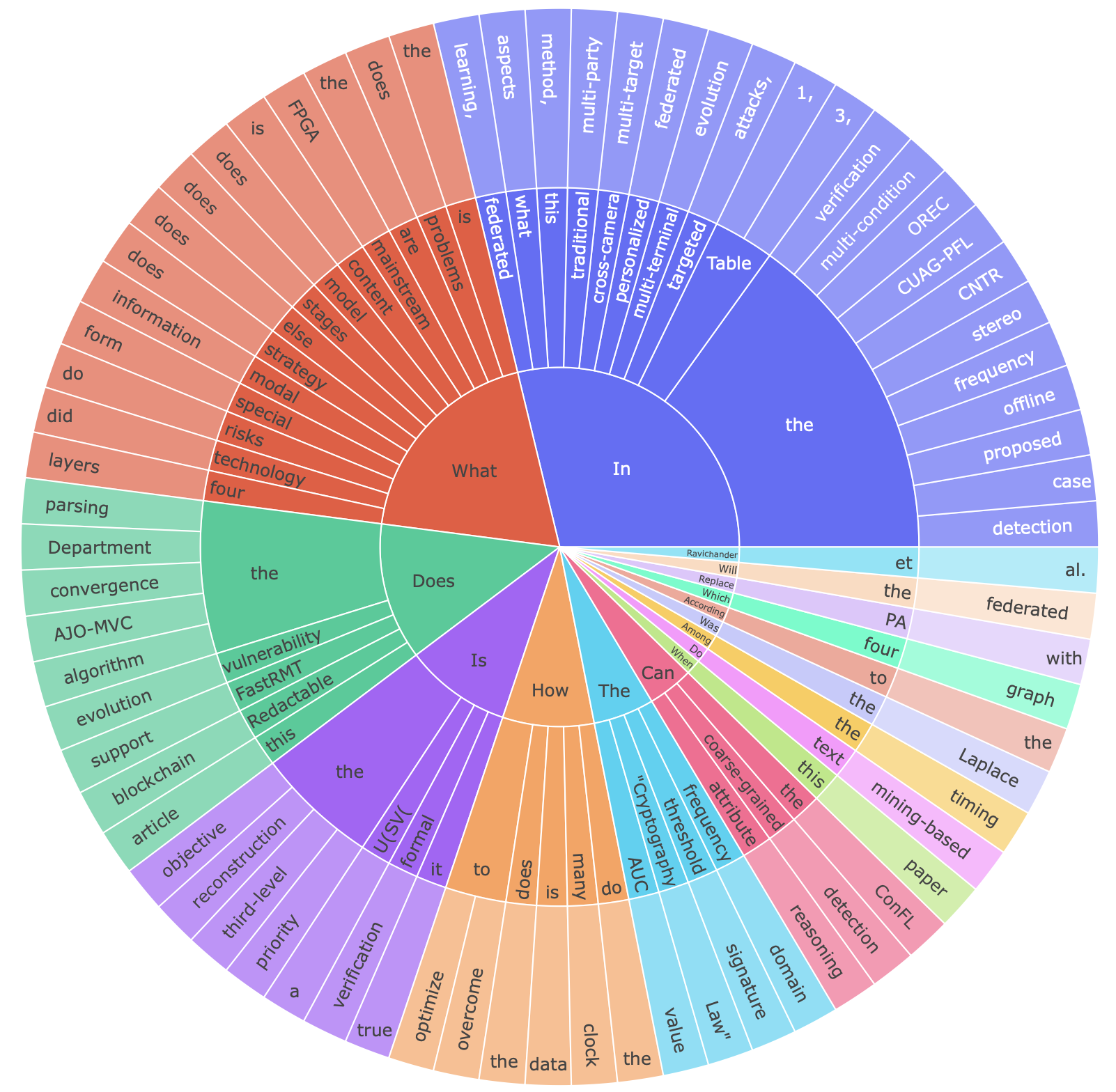}
        \caption{Chinese Paper}
    \end{subfigure}
    \caption{Data Visualization of XPaperQA, Chinese paper instruction was translated to English for clearer statistic.}
\end{figure*}

\clearpage

\section{XPaperQA construction prompt}

\subsection{Yes-no question generate prompt}
\begin{tcolorbox}[colback=blue!5!white,colframe=blue!75!black,title=Prompt For Yes-No (Chinese)]
\noindent\begin{CJK}{UTF8}{gbsn} 
你是一个科学文献问答对数据集处理专家。你的任务是根据我给出的文献内容，生成适合作为问答对数据集的问题和答案。问题要尽量学术。一句话中只能有一个问题。

你的问题需要是判断对错式的，对原文中的事实性知识做出判断或者判断需要推理的问题，答案只能是是或否。生成尽可能多的问题，问题要相似度低，答案一半为是，一半为否。你要针对你的回答生成置信度分数，分数区间为1至10的整数。

你必须按照我给出的问答对格式来生成：\{ "question": <your question here>, "answer": <your answer here>, "confidence": <your confidence score here> \}

请对特殊符号进行转义。我的文献内容如下：<my content here>
\end{CJK}
\end{tcolorbox}

\noindent\begin{tcolorbox}[colback=blue!5!white,colframe=blue!75!black,title=Prompt For Yes-No (English)]
You are an expert in processing scientific literature question-answering datasets. Your task is to generate questions and answers suitable for question-answering datasets based on the literature content I give you. Questions should be as academic as possible. There can only be one question in a sentence.

Your questions need to be true or false, and require judgments on factual knowledge in the original text or questions that require reasoning. The answers can only be yes or no. Generate as many questions as possible, with low similarity, and half of the answers are yes and half are no. You need to generate a confidence score for your answer, with the score range being an integer from 1 to 10.

You must generate it according to the question-answer format I gave:\{ "question": <your question here>, "answer": <your answer here>, "confidence": <your confidence score here> \}

Please escape special symbols. My document content is as follows: <my content here>
\end{tcolorbox}

\subsection{Extractive question generation prompt}

\begin{tcolorbox}[colback=blue!5!white,colframe=blue!75!black,title=Prompt For Extractive (Chinese)]
\begin{CJK}{UTF8}{gbsn} 
你是一个科学文献问答对数据集处理专家。你的任务是根据我给出的文献内容，生成适合作为问答对数据集的问题和答案。
问题要尽量学术。一句话中只能有一个问题。

生成的问题必须细节、有价值，不要生成特别宏观的问题。你的问题需要是抽取式的，即答案必须是文中出现的一个数字，词语或者短句。生成尽可能多的问题，问题要相似度低。你要针对你的回答生成置信度分数，分数区间为1至10的整数。

你必须按照我给出的问答对格式来生成：\{ "question": <your question here>, "answer": <your answer here>, "confidence": <your confidence score here> \}

请对特殊符号进行转义。我的文献内容如下：<my content here>
\end{CJK}
\end{tcolorbox}
\noindent\begin{tcolorbox}[colback=blue!5!white,colframe=blue!75!black,title=Prompt For Extractive (English)]
You are an expert in processing scientific literature question-answering datasets. Your task is to generate questions and answers suitable for question-answering datasets based on the literature content I give you. Questions should be as academic as possible. There can only be one question in a sentence.

The questions you generate must be detailed and valuable. Don't generate macro questions. Your questions need to be extractive, meaning the answer must be a number, word, or phrase that appears in the text. Generate as many questions as possible, with low similarity. You need to generate a confidence score for your answer, which is an integer between 1 and 10.

You must generate it according to the question-answer format I gave:\{ "question": <your question here>, "answer": <your answer here>, "confidence": <your confidence score here> \}

Please escape special symbols. My document content is as follows: <my content here>
\end{tcolorbox}

\subsection{Abstractive QA generation prompt}
\begin{tcolorbox}[colback=blue!5!white,colframe=blue!75!black,title=Prompt For Abstractive (Chinese)]
\begin{CJK}{UTF8}{gbsn} 
你是一个科学文献问答对数据集处理专家。你的任务是根据我给出的文献内容，生成适合作为问答对数据集的问题和答案。问题要尽量学术。一句话中只能有一个问题。

生成的问题要能概括捕捉文本中的要点和主旨,而不只是浅层次的细节。你的问题需要是摘要式的，即答案不是直接引用文章中的原文，而是需要一定推理、归纳、总结才能获得答案。生成尽可能多的问题，问题要相似度低。你要针对你的回答生成置信度分数，分数区间为1至10的整数。

你必须按照我给出的问答对格式来生成：\{ "question": <your question here>, "answer": <your answer here>, "confidence": <your confidence score here> \}

请对特殊符号进行转义。我的文献内容如下：<my content here>
\end{CJK}
\end{tcolorbox}
\begin{tcolorbox}[colback=blue!5!white,colframe=blue!75!black,title=Prompt For Abstractive (English)]
You are an expert in processing scientific literature question-answering datasets. Your task is to generate questions and answers suitable for question-answering datasets based on the literature content I give you. Questions should be as academic as possible. There can only be one question in a sentence.

The questions you generate should capture the main points and themes of the text, not just the superficial details. Your questions need to be summary-style, that is, the answers are not directly quoted from the original text of the article, but require some reasoning, induction, and summary to get the answer. Generate as many questions as possible, and the questions should have low similarity. You need to generate a confidence score for your answer, which is an integer between 1 and 10.

You must generate it according to the question-answer format I gave:\{ "question": <your question here>, "answer": <your answer here>, "confidence": <your confidence score here> \}

Please escape special symbols. My document content is as follows: <my content here>
\end{tcolorbox}
\subsection{Re-answer prompt}
\begin{tcolorbox}[colback=blue!5!white,colframe=blue!75!black,title=Re-answer prompt For Abstractive (Chinese)]
\begin{CJK}{UTF8}{gbsn} 
你是一个科学文献问答对数据集处理专家。你的任务是根据我给出的文献内容，以及我提出的问题，生成适合作为问答对数据集的答案，不要生成你的问题。问题比较学术。一句话中只能有一个问题。

我的问题能概括捕捉文本中的要点和主旨,而不只是浅层次的细节。我的问题需要答案是摘要式的，即答案不是直接引用文章中的原文，而是需要一定推理、归纳、总结才能获得答案。你要针对你的答案生成置信度分数，分数区间为1至10的整数。

你必须按照我给出的问答对格式来生成：\{ "question": <my question here>, "answer": <your answer here>, "confidence": <your confidence score here> \}

请对特殊符号进行转义。我的文献内容如下：<my content here>

我的问题如下：<my question here>
\end{CJK}
\end{tcolorbox}
\begin{tcolorbox}[colback=blue!5!white,colframe=blue!75!black,title=Re-answer prompt For Abstractive (English)]
You are an expert in processing scientific literature question-answering datasets. Your task is to generate answers suitable for question-answering datasets based on the literature content I give you and the questions I ask. Do not generate your own questions. The questions are more academic. There is only one question in one sentence.

My questions capture the main points and themes of the text, not just the superficial details. My questions require summaries, that is, the answers are not directly quoted from the original text, but require some reasoning, induction, and summary to get the answer. You need to generate a confidence score for your answer, with the score range being an integer from 1 to 10.

You must generate it according to the question-answer format I gave:\{ "question": <my question here>, "answer": <your answer here>, "confidence": <your confidence score here> \}

Please escape special symbols. My document content is as follows: <my content here>

My questions are as follows: <my question here> 
\end{tcolorbox}
\begin{tcolorbox}[colback=blue!5!white,colframe=blue!75!black,title=Re-answer prompt For Yes-No (Chinese)]
\begin{CJK}{UTF8}{gbsn} 
你是一个科学文献问答对数据集处理专家。你的任务是根据我给出的文献内容，以及我提出的问题，生成适合作为问答对数据集的答案，不要生成你的问题。问题比较学术。一句话中只有一个问题。

我的问题需要答案是判断对错式的，对原文中的事实性知识做出判断或者判断需要推理的问题，你的答案只能是是或否。你要针对你的答案生成置信度分数，分数区间为1至10的整数。

你必须按照我给出的问答对格式来生成：\{ "question": <my question here>, "answer": <your answer here>, "confidence": <your confidence score here> \}

请对特殊符号进行转义。我的文献内容如下：<my content here>

我的问题如下：<my question here>
\end{CJK}
\end{tcolorbox}
\begin{tcolorbox}[colback=blue!5!white,colframe=blue!75!black,title=Re-answer prompt For Yes-No (English)]
You are an expert in processing scientific literature question-answering datasets. Your task is to generate answers suitable for question-answering datasets based on the literature content I give you and the questions I ask. Do not generate your own questions. The questions are more academic. There is only one question in one sentence.

My questions require true or false answers. You need to make a judgment on the factual knowledge in the original text or make a judgment on the question that requires reasoning. Your answer can only be yes or no. You need to generate a confidence score for your answer, which ranges from 1 to 10.

You must generate it according to the question-answer format I gave:\{ "question": <my question here>, "answer": <your answer here>, "confidence": <your confidence score here> \}

Please escape special symbols. My document content is as follows: <my content here>

My questions are as follows: <my question here> 
\end{tcolorbox}

\noindent\begin{tcolorbox}[colback=blue!5!white,colframe=blue!75!black,title=Re-answer prompt For Extractive (Chinese)]
\begin{CJK}{UTF8}{gbsn} 
你是一个科学文献问答对数据集处理专家。你的任务是根据我给出的文献内容，以及我提出的问题，生成适合作为问答对数据集的答案，不要生成你的问题。问题比较学术。一句话中只能有一个问题。

我的问题是细节、有价值的问题，不是特别宏观的问题。我的问题需要答案是抽取式的，即答案必须是文中出现的一个数字，词语或者短句。你要针对你的答案生成置信度分数，分数区间为1至10的整数。

你必须按照我给出的问答对格式来生成：\{ "question": <my question here>, "answer": <your answer here>, "confidence": <your confidence score here> \}

请对特殊符号进行转义。我的文献内容如下：<my content here>

我的问题如下：<my question here>
\end{CJK}
\end{tcolorbox}
\begin{tcolorbox}[colback=blue!5!white,colframe=blue!75!black,title=Re-answer prompt For Extractive (English)]
You are an expert in processing scientific literature question-answering datasets. Your task is to generate answers suitable for question-answering datasets based on the literature content I give you and the questions I ask. Do not generate your own questions. The questions are more academic. There is only one question in one sentence.

My questions are detailed and valuable, not particularly macro questions. My questions require extractive answers, that is, the answer must be a number, word or short sentence that appears in the text. You need to generate a confidence score for your answer, which is an integer between 1 and 10.

You must generate it according to the question-answer format I gave:\{ "question": <my question here>, "answer": <your answer here>, "confidence": <your confidence score here> \}

Please escape special symbols. My document content is as follows: <my content here>

My questions are as follows: <my question here> 
\end{tcolorbox}
\subsection{Consistency filter prompt}
\noindent\begin{tcolorbox}[colback=blue!5!white,colframe=blue!75!black,title=Prompt For Consistence Filter (Chinese)]
\begin{CJK}{UTF8}{gbsn} 
你是一个科学文献问答对数据集处理专家。你的任务是根据我给出的两个问答，判断两个同样的问题对应的回答是否基本一致，答案只能是一致或不一致。你要针对你的答案生成一致度分数，分数区间为 1 至 10 的整数。

你必须按照我给出的格式来生成：\{ "question": <my question here>, "answer": <your answer here>, "score": <your consistent score here> \}

请对特殊符号进行转义。我的问答内容如下：\{ "question": <my question here>, "answer": <my answer here>, "confidence": <my confidence score here> \} \{ "question": <my question here>, "answer": <my answer here>, "confidence": <my confidence score here> \}
\end{CJK}
\end{tcolorbox}
\begin{tcolorbox}[colback=blue!5!white,colframe=blue!75!black,title=Prompt For Consistence Filter (English)]
You are an expert in processing scientific literature question-answering datasets. Your task is to determine whether the answers to the two same questions I gave are basically consistent, and the answers can only be consistent or inconsistent. You need to generate a consistency score for your answer, with the score range being an integer from 1 to 10.

You must generate it according to the question-answer format I gave:\{ "question": <my question here>, "answer": <your answer here>, "score": <your consistent score here> \}

Please escape special symbols. My question and answer are as follows:\{ "question": <my question here>, "answer": <my answer here>, "confidence": <my confidence score here> \} \{ "question": <my question here>, "answer": <my answer here>, "confidence": <my confidence score here> \}
\end{tcolorbox}

\end{document}